\pdfoutput=1
\documentclass[11pt]{article}

\usepackage[preprint]{acl}

\usepackage{times}
\usepackage{latexsym}

\usepackage[T1]{fontenc}

\usepackage[utf8]{inputenc}

\usepackage{microtype}

\usepackage{inconsolata}

\usepackage{graphicx}

\usepackage{glossaries}
\usepackage{amsmath,amsfonts,amssymb}
\usepackage[ruled,vlined,noend,noline]{algorithm2e}
\usepackage{multirow}
\usepackage{graphicx}
\usepackage[normalem]{ulem}
\useunder{\uline}{\ul}{}
\usepackage{colortbl}
\usepackage{hhline}
\usepackage{tcolorbox}
\usepackage{float}
\usepackage{fancyvrb}
\usepackage{placeins}
\usepackage{amsthm}

\glsdisablehyper
\newacronym{llm}{LLM}{large language model}
\newacronym{nlp}{NLP}{natural language processing}
\newacronym{rag}{RAG}{retrieval-augmented generation}
\newacronym{mas}{MAS}{multi-agent system}
\newacronym{mass}{MASS}{multi-agent search system}
\newacronym{sft}{SFT}{Supervised Fine-tuning}
\newacronym{rl}{RL}{Reinforcement Learning}
\newacronym{marl}{MARL}{Multi-Agent Reinforcement Learning}
\newacronym{dpo}{DPO}{Direct Preference Optimization}
\newacronym{ppo}{PPO}{Proximal Policy Optimization}
\newacronym{gae}{GAE}{Generalized Advantage Estimation}
\newacronym{goa}{GOA}{Group-based Optimization Algorithm}
\newacronym{grpo}{GRPO}{Group Relative Policy Optimization}
\newacronym{dapo}{DAPO}{Decoupled
Clip and Dynamic sAmpling Policy Optimization}

\newacronym{mappo}{MAPPO}{Multi-Agent Proximal Policy Optimization}

\newacronym{em}{EM}{Exact Match}

\newacronym{is}{IS}{Independent Sampling}
\newacronym{fof}{FoF}{Fork-on-first}
\newacronym{cp}{CP}{Cherry-pick}
\newacronym{rr}{RR}{Round-robin}

\newacronym{ours}{MHGPO}{Multi-Agent Heterogeneous Group Policy Optimization}

\title{End-to-End Optimization of LLM-Driven Multi-Agent Search Systems via Heterogeneous-Group-Based Reinforcement Learning}

\author{
  Guanzhong Chen$^{1}$\thanks{~~Work done during internship.} \thanks{~~Equal contribution.} \quad
  Shaoxiong Yang$^{1}$\footnotemark[2] \quad
  Chao Li$^{1}$\thanks{~~Corresponding author.} \\
  \textbf{Wei Liu}$^{1}$ \quad
  \textbf{Jian Luan}$^{1}$ \quad
  \textbf{Zenglin Xu}$^{2,3}$\footnotemark[3] \\
  $^1$MiLM Plus, Xiaomi Inc. \; $^2$Fudan University \; $^3$Shanghai Academy of AI for Science \\
  \texttt{muxichenz@outlook.com} \\
  \texttt{\{yangshaoxiong, lichao75, liuwei40, luanjian\}@xiaomi.com} \\
  \texttt{zenglinxu@fudan.edu.cn}
}

\begin{document}
\maketitle

\begin{abstract}
Large language models (LLMs) are versatile, yet their deployment in complex real-world settings is limited by static knowledge cutoffs and the difficulty of producing controllable behavior within a single inference. Multi-agent search systems (MASS), which coordinate specialized LLM agents equipped with search tools, mitigate these issues via task decomposition and retrieval-augmented problem solving. However, optimizing LLMs for agent-specific roles remains labor-intensive with prompt engineering or supervised fine-tuning, motivating automated end-to-end training. Existing multi-agent reinforcement learning (MARL) methods such as Multi-Agent Proximal Policy Optimization (MAPPO) typically depend on large critic networks to evaluate joint actions, leading to instability and high memory costs. We introduce Multi-Agent Heterogeneous Group Policy Optimization (MHGPO), which updates policies by estimating relative advantages across heterogeneous groups of multi-agent rollouts, shifting the optimization focus from local agent performance to global system success. We further study three group rollout sampling strategies to trade off sample efficiency and optimization quality. Experiments show that MHGPO captures implicit inter-agent dependencies and consistently outperforms strong baselines in both task performance and computational efficiency.
\end{abstract}
\section{Introduction}

\Glspl{llm} have demonstrated strong performance across a wide range of domains. Despite these advances, deploying \glspl{llm} in real-world industrial settings remains challenging. On the one hand, the fixed knowledge cutoff of \glspl{llm}~\citep{rag-survey} limits their ability to reason over unseen information. On the other hand, although advanced reasoning models such as DeepSeek-R1~\citep{DBLP:journals/nature/GuoYZSWZXZMBZY025} have achieved substantial progress on complex tasks, \glspl{llm} still struggle to produce controllable and consistently accurate outputs within a single inference.

As a result, practical \gls{llm} deployment in specialized scenarios increasingly relies on combining \emph{agent-based} frameworks with \gls{rag}. In contrast to methods that augment a single strong reasoning model with retrieval inside a \emph{single-context} reasoning trace~\citep{DBLP:journals/corr/abs-2501-05366, DBLP:journals/corr/abs-2503-05592}, a widely adopted alternative is to organize a team of specialized \gls{llm} agents as an \gls{mas}, where each agent plays a well-defined role, is equipped with \emph{search tools}, and coordinates with others through structured communication. This decomposes high-level objectives into modular, \emph{multi-context} subtasks, reducing the burden on any single model while improving system interpretability and controllability. We refer to such systems---\glspl{mas} augmented with \gls{rag} via search tools---as \glspl{mass}, the setting studied in this work.

Numerous studies have explored \glspl{mass} for complex functionalities~\citep{DBLP:journals/corr/abs-2501-00332, DBLP:journals/corr/abs-2501-15228, DBLP:journals/corr/abs-2401-06800, DBLP:journals/corr/abs-2501-07813, DBLP:journals/corr/abs-2503-06951, DBLP:conf/emnlp/ChenZWWSCX25}. Yet in practice \glspl{llm} often fail to follow instructions or use tools effectively, e.g., issuing retrieval queries misaligned with the search engine. Prompt engineering and per-agent \gls{sft} can mitigate these issues but incur substantial engineering overhead and adapt poorly to evolving requirements, motivating \gls{rl}-based methods for robust, \emph{end-to-end} system-level optimization.

\gls{rl} methods for \glspl{llm} enhance capabilities by rewarding desirable behaviors and driving policy exploration toward higher expected return. In \glspl{mas} where \glspl{llm} underpin individual agents, \gls{rl} enables end-to-end optimization of both agent competence and overall system performance, typically cast as \gls{marl}. By assigning rewards to system-level outcomes and fine-tuning the underlying \glspl{llm}, \gls{marl} supports task-adaptive refinement within the \gls{mas}. However, \gls{marl} faces two central challenges: \emph{(i)} efficiently optimizing diverse agents toward global optima with minimal overhead, and \emph{(ii)} properly attributing system-level outcomes to individual agents whose individual contributions are not directly measurable from the global reward.

Recent work applies \gls{mappo}~\citep{DBLP:journals/corr/SchulmanWDRK17, DBLP:conf/nips/YuVVGWBW22} to optimize \glspl{mas}~\citep{liao2025marftmultiagentreinforcementfinetuning, DBLP:journals/corr/abs-2501-15228} in an actor--critic fashion, with the \gls{llm} as the \emph{actor} and a large \emph{critic} estimating returns. However, approximating joint action values over diverse agents is often unstable and incurs substantial memory and compute overhead, limiting scalability. By contrast, recent \glspl{goa} such as \gls{grpo}~\citep{DBLP:journals/corr/abs-2402-03300, DBLP:journals/corr/abs-2503-14476} remove the critic and estimate advantages from relative rewards across group rollouts. While effective in \emph{single-context} settings, extending them to \emph{multi-context} \glspl{mass} is nontrivial: a multi-agent rollout spans several agents with disjoint local contexts, so optimization must account for \emph{(i)} \textbf{indirect inter-agent dependencies}, where an upstream agent's output shapes downstream behavior without a direct gradient path, and \emph{(ii)} \textbf{implicit cross-trajectory relations}, where rollouts from the same root query explore related but non-identical intermediate decisions and jointly reflect system-level success. Local, prefix-conditioned comparisons alone are therefore insufficient.

To bridge this gap, we present, to our knowledge, the first systematic study of \glspl{goa} for \gls{mass} and introduce \gls{ours}, a \gls{goa} that optimizes \glspl{mass} via multi-agent rollout sampling and backward reward propagation, with advantages estimated from relative rewards within \emph{heterogeneous groups}. Our main contributions are:

\begin{itemize}
\item Leveraging \textbf{parameter sharing} and \textbf{critic-free training}, \gls{ours} recasts \gls{marl} as \emph{multi-task} optimization, achieving cheaper and more stable learning than \acrshort{ppo}-based methods.

\item \gls{ours} combines \textbf{backward reward propagation} to capture \emph{indirect} inter-agent dependencies with \textbf{heterogeneous-group advantage estimation} to model \emph{implicit} cross-trajectory correlations, shifting optimization from local agent performance toward \emph{global system success}.

\item We explore three rollout sampling strategies for \glspl{mass}---\gls{is}, \gls{fof}, and \gls{rr}---and their efficiency--effectiveness trade-offs.

\item We formalize the first connection between \gls{ours} and single-context \gls{grpo}, showing that \gls{ours} yields a \textbf{near-aligned per-token advantage} under an idealized context-sufficiency assumption yet empirically excels in stability and efficiency via multi-context optimization.

\item On mainstream benchmarks, \gls{ours} consistently outperforms strong baselines in both task performance and computational efficiency.
\end{itemize}
\section{Preliminary}

\subsection{\gls{ppo}}
\begin{figure}
    \centering
    \includegraphics[width=\linewidth]{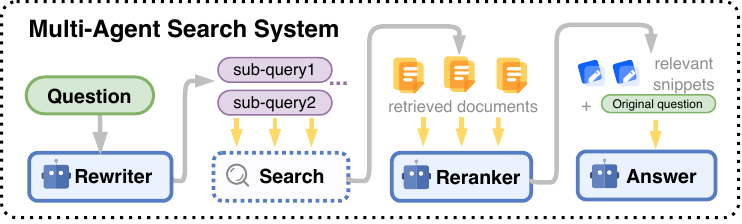}
    \caption{Illustration of the three-agent \gls{mass}: the \emph{Rewriter} decomposes the question into sub-queries, the \emph{Reranker} selects relevant snippets from retrieved documents, and the \emph{Answerer} generates the final answer.}
    \label{fig:mas}
\end{figure}

\begin{figure*}
    \centering
    \includegraphics[width=1\textwidth]{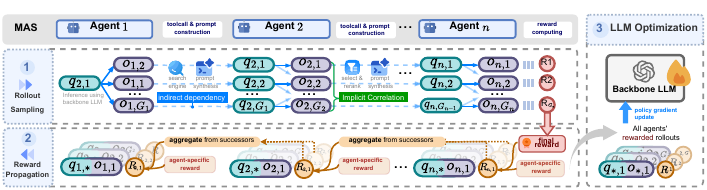}
    \caption{The proposed \gls{ours} framework proceeds as follows. For each \gls{mas} input (question), we first sample multi-agent rollouts, producing multiple trajectories. Each trajectory receives a terminal reward, which is propagated backward and aggregated to assign a reward to each rollout across trajectories. The resulting reward-labeled rollouts are then used to optimize the backbone \gls{llm} with Equation~\ref{eq:target}.}
    \label{fig:marl}
\end{figure*}

When adapted to \gls{llm} optimization, \gls{ppo} retains the standard actor--critic structure with three models: an actor (policy) $\pi_{\theta}$, a critic $V_{\phi}$, and a frozen reference policy $\pi_{\theta_{\text{ref}}}$. Given a prompt $q$, the actor samples an output $o \sim \pi_{\theta}(\cdot \mid q)$. The critic estimates $V_{\phi}(q,o)$, a scalar baseline for advantage computation. The reference policy regularizes updates (typically via a KL penalty), stabilizing training. \gls{ppo} then maximizes the following objective:

\begin{equation}
\begin{aligned}
\mathcal{J}_{\text{PPO}}(\theta) &= {\mathbb{E}}_{q\sim \mathcal{D},o_{\leq t}\sim \pi_{\theta_{\text{old}}}(\cdot|q)} \\ &[\min(r_t\hat{A_t}, \text{clip}(r_t,1-\epsilon, 1+\epsilon)\hat{A}_t)], 
\end{aligned}
\end{equation} 
where 
$r_t=\frac{\pi_{\theta}(o_t \mid q, o_{<t})}{\pi_{\theta_{\text{old}}}(o_t \mid q, o_{<t})}$ 
is the importance-sampling ratio that constrains the policy update magnitude, and $\hat{A}_t$ denotes the advantage at time step $t$.  
In \gls{ppo}, $\hat{A}_t$ is typically computed using \gls{gae}.  
For the $l$-th token of the generated response $o$, the critic provides a value estimate $V_l$, while the reward model (or scoring function) assigns a reward $R_l$, with $l=0,1,2,\ldots$.  
The advantage is estimated as
\begin{equation}
\hat{A}_t = \sum_{l=0}^{\infty} (\gamma \lambda)^l \delta_{t+l}, 
\qquad
\delta_l = R_l + \gamma V_{l+1} - V_l,
\end{equation}
where $\gamma \in [0,1]$ and $\lambda \in [0,1]$ are the discount and GAE hyperparameters, respectively.


\subsection{\acrlong{grpo}}

Compared with \gls{ppo}, \gls{grpo} dispenses with an explicit critic and instead computes advantages directly from within-group relative rewards. Specifically, for an input question $q$, \gls{grpo} samples a group of $G$ responses $\{o_i\}_{i = 1}^G$. The advantage of the $i$-th response (shared across all time steps or applied only to the last time step) is then calculated as:

\begin{equation}
\label{eq:grpo-adv}
\hat{A}_{i} = \frac{R_i - \text{mean}(\{R_i\}_{i = 1}^G)}{\text{std}(\{R_i\}_{i = 1}^G)}
\end{equation} 



\subsection{Multi-Agent Search System}
In this paper, we utilize a simple yet effective three-agent \gls{mass} as an example to investigate \gls{marl} algorithms, as illustrated in Figure~\ref{fig:mas}. Following the design of previous work~\citep{DBLP:journals/corr/abs-2501-15228, DBLP:journals/corr/abs-2305-14283, DBLP:journals/corr/abs-2501-00332}, this system comprises a \emph{Rewriter}, responsible for generating retrieval queries tailored for search engines based on the original questions; a \emph{Reranker}, which selects relevant items from a large pool of retrieval results to aid in answering the original questions; and an \emph{Answerer}, which produces the final answers by integrating the original questions with the filtered and reranked retrieval results. This system can invoke external retrieval tools to address complex queries by leveraging supplementary information.

\section{Method: \gls{ours}}

In this section, we present \acrfull{ours}, the first \gls{goa} for \glspl{mas} that estimates relative advantages over \emph{heterogeneous} rollout groups. \gls{ours} comprises a general policy optimization framework and three group-based rollout sampling strategies. Although we focus on \glspl{mass} as the primary setting, the formulation applies more broadly to general \glspl{mas}; accordingly, we use \gls{mass} as the running example while keeping the notation general.

\subsection{Multi-Agent Policy Optimization Framework}

Following \gls{mappo}~\citep{DBLP:conf/nips/YuVVGWBW22}, \gls{ours} uses \emph{parameter sharing}: all $n$ agents ($k\in{1,\ldots,n}$) are instantiated with a single \gls{llm} backbone, which is jointly optimized to perform each agent role. This design simplifies training and reduces compute and memory overhead. Figure~\ref{fig:marl} summarizes the overall \gls{rl} framework; details follow.

\paragraph{Multi-Agent Group Rollout Sampling.}  Given a single question $q \sim D$ sampled from the \gls{rl} dataset, the \gls{mas} performs \emph{group rollout sampling} by invoking its internal agents to collaboratively generate $G$ final responses $\{o_{i}\}_{i=1}^{G}$. The sequence of intermediate steps from the input $q$ to each final response constitutes a rollout \emph{trajectory}. During sampling, the $k$-th agent produces a total of $G_k$ input-output pairs, each corresponding to a distinct trajectory, denoted as $\{(q_{k,i}, o_{k,i}, m_{k,i})\}_{i = 1}^{G_k}$. Here, $m_{k,i}$ denotes the \emph{group identifier} associated with each rollout, as determined by a specific group rollout sampling algorithm. This identifier is subsequently used to aggregate rollouts into coherent groups. The resulting input-output pairs are collected as \emph{agent rollouts} and serve as training data for model optimization.


\paragraph{Backward Reward Propagation: Capturing Indirect Inter-Agent Dependencies} After the \gls{mas} generates the final responses $\{o_{i}\}_{i=1}^{G}$ for a query $q$, a reward model or predefined reward rule assigns reward signals to these outputs, resulting in a shared reward set $\{R^{\text{shared}}_{i}\}_{i=1}^{G}$. Starting from the endpoints of the sampling trajectories, these shared rewards are then \emph{propagated backward} through each trajectory to reach the preceding agents $k$, where they are \emph{aggregated} for each agent's output. Specifically, for the $i$-th output $o_{k,i}$ produced by agent $k$, the corresponding shared reward is computed as:
\begin{equation}
\label{eq:aggr}
    R^{\text{shared}}_{k,i} = \text{Aggr}(\{R_{j,r}\}_{j > k})
\end{equation}
where $j$ is the \emph{direct successor} of agent $k$ along the trajectory---i.e., $j$ consumes $k$'s output with no intermediate agents. We call $(k,j)$ an \emph{indirect-dependent} pair. Each $o_{j,r}$ is a response generated by agent $j$ from inputs that include $o_{k,i}$. $\mathrm{Aggr}(\cdot)$ denotes an aggregation operator, implemented as simple averaging by default. This reward-sharing scheme exposes indirect upstream--downstream dependencies even when agents do not share context.

After the shared rewards have been propagated, the final reward for each output $\{o_{k,i}\}_{i=1}^{G_k}$ is calculated by incorporating the agent-specific reward function $R_k^{\text{spe}}(\cdot)$, which typically imposes a penalty based on the output format of the agent. The resulting reward is given by $R_{k,i} = R_{k,i}^{\text{shared}} + R_k^{\text{spe}}(q_{k,i}, o_{k,i})$. Concretely, in our three-agent \gls{mass} (Rewriter $\rightarrow$ Reranker $\rightarrow$ Answerer), the Answerer's F1 against the gold answer serves as the terminal shared reward, which is then propagated backward and averaged to the Reranker, and further to the Rewriter.

\paragraph{Heterogeneous Group Advantage Estimation: Implicit Cross-Trajectory Correlation Modeling}
Following \glspl{goa}, we estimate advantages using group-wise relative rewards, eliminating the need for a critic. For a single system-level input, the advantage of the $i$-th rollout from agent $k$ is
\begin{equation}
\label{eq:adv}
\hat{A}_{k,i} =
\frac{
R_{k,i} - \mathrm{mean}\!\left(\{R_{l,j}\mid m_{l,j}=m_{k,i}\}\right)
}{
\mathrm{std}\!\left(\{R_{l,j}\mid m_{l,j}=m_{k,i}\}\right)
}.
\end{equation}
Here, advantages are normalized within the set of samples sharing the same \emph{group identifier} $m$. The resulting scalar advantage is broadcast to all tokens, i.e., $A_{k,i}^t = A_{k,i}$. Unlike single-agent \gls{grpo}, a group may include rollouts from different prompts, yielding \emph{heterogeneous groups}. This cross-trajectory normalization operationalizes the \emph{implicit cross-trajectory relations} introduced in the introduction: by comparing rollouts that stem from the same root query but differ in intermediate decisions (e.g., different Rewriter sub-queries, such as asking about \emph{birthplace} vs.\ \emph{nationality} for the same multi-hop question), the advantage signal no longer merely selects the best local action under a fixed upstream prefix, but instead rewards rollouts that lead to \emph{globally successful} system behavior. In effect, heterogeneous grouping intentionally shifts the optimization focus from local agent performance to global system success, at the cost of additional variance in the relative-reward signal---a trade-off we further address through our \gls{rr} sampling strategy introduced below.


\paragraph{Multi-Agent Optimization}  
Under parameter sharing, the \gls{rl} objective for the \gls{mas} takes the same form as \gls{grpo}; however, training requires \emph{aggregating} the losses from the $n$ agents to enable collaborative optimization---casting multi-agent learning as a multi-task learning problem.
\begin{equation}
\label{eq:target}
\begin{aligned}&\mathcal{J}_{\text{\tiny\gls{ours}}}(\theta) = \mathbb{E}_{\substack{\{o\} \sim {\text{\tiny$\mathbf{MAS}$}}_{\theta_\text{old}} (\cdot|q)\\ q\sim \mathcal{D}}} \Big[ \frac{1}{n}\boldsymbol{\sum_{k=1}^n} \frac{1}{G_k}\sum_{i = 1}^{G_k}\frac{1}{|o_{k,i}|}\\ &\sum_{t = 1}^{|o_{k,i}|}\min\!\big(r^t_{k,i}\hat{A}^t_{k,i}, \text{clip}(r^t_{k,i},1-\epsilon, 1+\epsilon)\hat{A}^t_{k,i}\big) \\ &-\beta\, D_{\mathrm{KL}}\!\left(\pi_\theta\,\|\,\pi_{\text{ref}}\right)\Big],
\end{aligned}
\end{equation}
\begin{equation}
r^t_{k,i}=\frac{\pi_{\theta}(o^t_{k,i}|q_{k,i},o_{k,i}^{<t})}{\pi_{\theta_{\text{old}}}(o_{k,i}^t|q_{k,i},o_{k,i}^{<t})}.
\end{equation}
With the above objective, coordinated optimization of the multi-agent system can be performed based on all internal input-output pairs generated during the \gls{mas} rollout phase.


\subsection{Rollout Sampling Strategies}

A key challenge in the proposed multi-agent optimization framework is constructing the candidate group $G_k'$ for each agent: executing \emph{group rollouts} to obtain $G_k$, and then \emph{forming completion groups} from the sampled trajectories. This section explores three strategies in \gls{ours}, each using a different mechanism for rollout sampling and group formation (Figures~\ref{fig:rollout-bf} and~\ref{fig:rollout}); full algorithms are given in the Appendix.


\begin{figure}
    \centering
    \includegraphics[width=\linewidth]{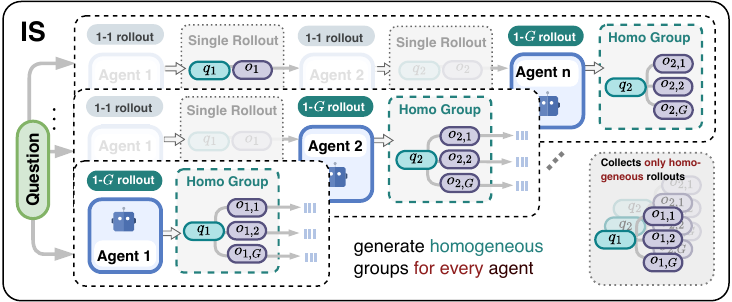}
    \caption{The all-homogeneous rollout strategy \emph{\acrfull{is}}, where each sample is propagated to all agents for one-to-$G$ rollouts. Only the resulting homogeneous groups are used for model updates.}
    \label{fig:rollout-bf}
\end{figure}

\begin{figure}
    \centering
    \includegraphics[width=\linewidth]{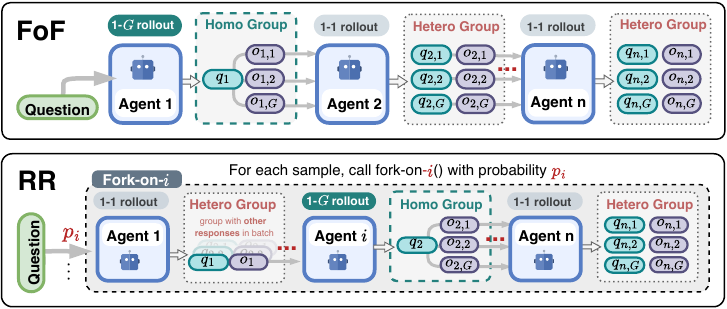}
    \caption{The proposed two \emph{heterogeneous-group-based} rollout sampling strategies, namely \emph{\acrfull{fof}} and \emph{\acrfull{rr}}.}
    \label{fig:rollout}
\end{figure}

\paragraph{All Homogeneous: \acrfull{is}} We first introduce a baseline strategy for group rollout sampling, \gls{is} (Figure~\ref{fig:rollout-bf}). Following the relative-advantage formulation in \gls{grpo} (Equation~\ref{eq:grpo-adv}), each group is formed by sampling multiple rollouts from the \emph{same} input, yielding a \emph{homogeneous group}. Computing relative advantages within such groups provides a consistent, input-conditioned comparison of the \gls{llm}'s responses.

Specifically, for each question $q \sim D$, we sequentially run the \gls{mas} to each agent and, upon reaching that agent, sample a 1-$G$ group of rollouts. Rewards are computed via the backpropagation mechanism in \gls{ours}. Repeating this for all agents yields only 1-$G$ homogeneous rollouts, totaling $n \times G$ samples organized into $n$ independent homogeneous groups of size $G$. These are then used to fine-tune the backbone \gls{llm}, encouraging balanced performance across agent-specific tasks.

\paragraph{Introduce Heterogeneity: \acrfull{fof}} The homogeneous-only \gls{is} baseline is inefficient and myopic: it incurs substantial redundant rollouts and optimizes each agent independently, ignoring inter-agent interactions. We therefore propose \gls{fof}, a simple alternative that removes redundant sampling while explicitly capturing interactions among agents (Figure~\ref{fig:rollout}, upper). Crucially, \gls{fof} supports advantage estimation over \emph{heterogeneous groups}.

For each question, \gls{fof} samples $G$ rollouts \emph{only} at the entry agent of the \gls{mas}; all subsequent agents perform one-to-one rollouts. Consequently, the entry agent produces $G$ rollouts from the same input (a homogeneous group), while each downstream agent receives $G$ different inputs induced by upstream rollouts (forming \emph{heterogeneous groups}). Relative advantages are computed within the corresponding groups, producing $G$ reward signals at the system output and requiring only $nG$ rollout operations over the entire \gls{mas}.

\paragraph{Less Utilization for More Homogeneity: \acrfull{rr}}
\gls{fof} improves per-agent sample utilization without discarding rollouts and performs well in our preliminary results (Figure~\ref{fig:training}). However, because it always forks at the first agent, only the entry agent forms homogeneous rollout groups; downstream agents rarely receive identical inputs when estimating advantages, which can slow convergence. We therefore propose \gls{rr}, which randomizes the fork point across agents to serve as a bridge between global coordination pressure and local learning stability. For each input sample, we select agent $i$ as the fork point with probability $p_i$. If the sample forks at $i$, agents $j \ge i$ follow the \gls{fof} procedure for advantage estimation, while agents $j < i$ execute only one rollout. For each $j < i$, we compute relative advantages by grouping that rollout with $\left\lfloor p_i \cdot \text{batch\_size} \right\rfloor - 1$ other samples in the same batch that also fork at $i$. Consequently, each agent forms homogeneous rollout groups with probability $p_i$, while only $p_1 \cdot \text{batch\_size}$ samples incur the full \gls{fof} rollout cost; the rest require fewer rollouts, reducing overall resource usage.

\paragraph{More Utilization: \gls{fof} with Over-sampling}
Complementary to \gls{rr}, we also study whether higher sample utilization improves optimization under the \gls{fof} framework, despite increased training cost. Concretely, for each sampled trajectory we over-sample the \emph{terminal agent} by generating $G$ rollouts (instead of one), yielding a denser estimate of the final reward. We then propagate and aggregate this signal to earlier agents for reward estimation. We denote this variant as \emph{\gls{fof}(os)}.


\section{Experiments \& Discussion}

\subsection{Experiment Setup}

\paragraph{Implementation Details}
We instantiate each agent using prompts adapted from \citet{DBLP:journals/corr/abs-2501-15228}.
Following \citet{DBLP:conf/acl/SuTA0024}, we adopt a Wikipedia dump as the retrieval corpus and deploy \texttt{contriever}~\citep{contriever} as the search backend for the \emph{retriever} module.
Consistent with standard multi-hop RL practice, we train for one epoch.
Unless stated otherwise, \gls{ours} uses group rollouts with size $G=4$, and \gls{ours}-\gls{rr} uses round-robin probabilities $(0.7, 0.1, 0.2)$. Appendix Table~\ref{tab:group} shows that group sizes 4--6 perform similarly; we choose $G=4$ as the default because it offers near-best performance with lower rollout cost than larger groups.
Further details, including the effect of varying $G$, are provided in the Appendix.

\paragraph{Datasets and Models} We adopt commonly used evaluation datasets in search systems, including the multi-hop datasets HotpotQA~\citep{hotpot}, 2WikiMultihopQA~\citep{2wiki}, and MuSiQue~\citep{musique}, for training and evaluation. We use Llama3.1-8B-Instruct as the shared backbone of the \gls{mas}.

\paragraph{Evaluation Metrics} Since all datasets provide gold-standard answers, we report three answer-level metrics: Accuracy, \gls{em}, and F1. Accuracy measures token-level prediction accuracy between the prediction and the gold answer. \gls{em} requires an exact normalized match, while F1 measures token-level overlap.

\subsection{Main Results} 
\begin{figure}
    \centering
    \includegraphics[width=1\linewidth]{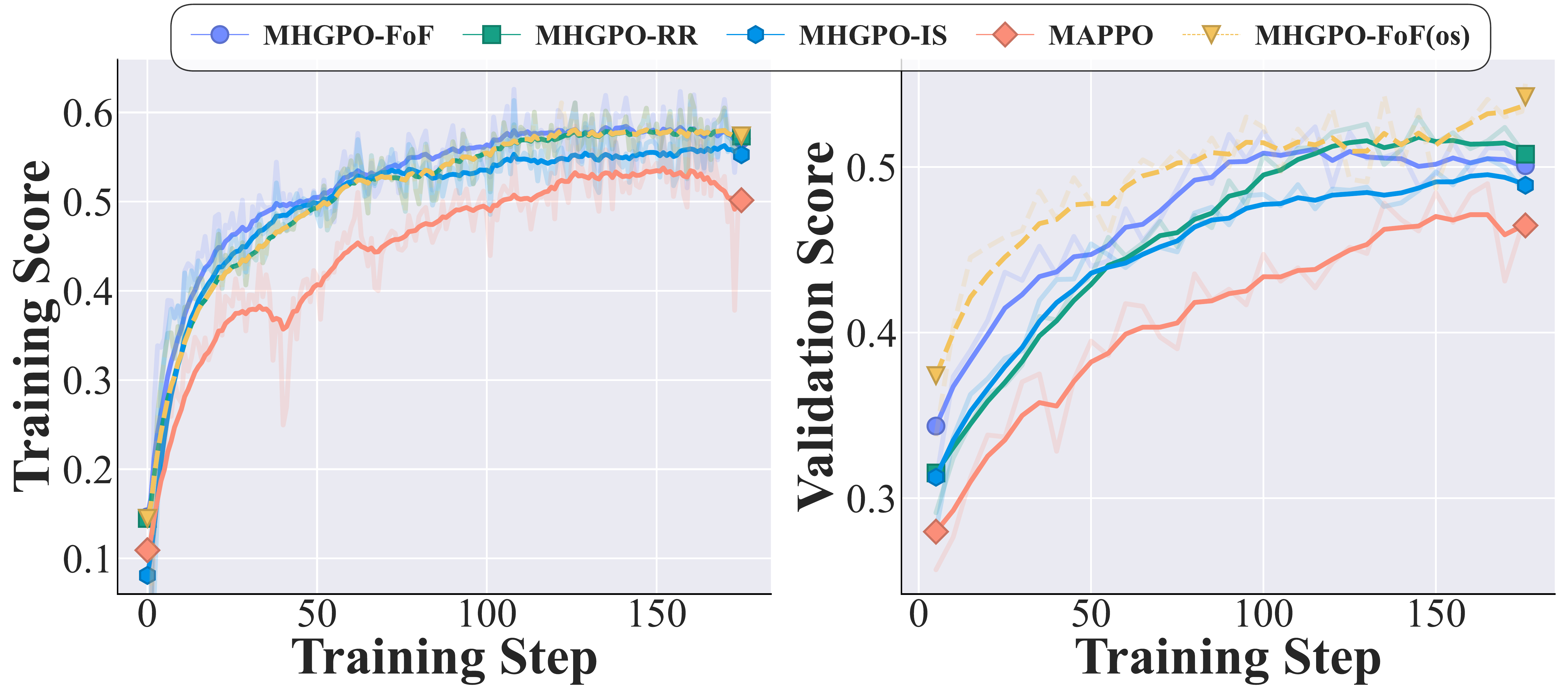}
    \caption{Training reward and validation F1 on a fixed 500-example HotpotQA subset during one epoch of training with different \gls{rl} algorithms. The validation subset is evaluated every 5 steps.}
    \label{fig:training}
\end{figure}

\begin{table*}[]
\caption{Test performance of \gls{rl} algorithms and training-free baselines on mainstream QA datasets, measured by Accuracy (\%; normalized answer-hit), Exact Match (\%), and F1 (\%). All trainable methods fine-tune the same Llama3.1-8B-Instruct checkpoint with one epoch of \gls{rl} on HotpotQA; $\ast$ denotes out-of-domain (OOD) evaluation.}
\label{tab:main}
\resizebox{\textwidth}{!}{%
\renewcommand{\arraystretch}{1.4}
\begin{tabular}{ll|lll|lll|lll|
>{\columncolor[HTML]{EFEFEF}}r 
>{\columncolor[HTML]{EFEFEF}}r }
\hline
\multicolumn{2}{l|}{}                                                                                                 & \multicolumn{3}{c|}{\textbf{HotpotQA}}                                     & \multicolumn{3}{c|}{*2WikiMultihopQA}                                      & \multicolumn{3}{c|}{*MuSiQue}                                              & \multicolumn{2}{c}{\cellcolor[HTML]{EFEFEF}AVG}                                                  \\
\multicolumn{2}{l|}{\multirow{-2}{*}{Methods/Models}}                                                                 & \multicolumn{1}{c}{Acc} & \multicolumn{1}{c}{EM} & \multicolumn{1}{c|}{F1} & \multicolumn{1}{c}{Acc} & \multicolumn{1}{c}{EM} & \multicolumn{1}{c|}{F1} & \multicolumn{1}{c}{Acc} & \multicolumn{1}{c}{EM} & \multicolumn{1}{c|}{F1} & \multicolumn{1}{c}{\cellcolor[HTML]{EFEFEF}ID} & \multicolumn{1}{c}{\cellcolor[HTML]{EFEFEF}OOD} \\ \hline
\multicolumn{1}{l|}{}                                                                            & Llama3.1-8B        & 21.269                  & 14.018                 & 22.782                  & 16.963                  & 7.069                  & 20.817                  & 5.492                   & 1.365                  & 2.813                   & 19.356                                         & 9.087                                           \\
\multicolumn{1}{l|}{\multirow{-2}{*}{LLM}}                                                       & Qwen2.5-72B        & 30.304                  & 25.564                 & 36.183                  & 29.373                  & 27.234                 & 32.038                  & 7.737                   & 5.337                  & 15.411                  & 30.684                                         & 19.522                                          \\ \hline
\multicolumn{1}{l|}{}                                                                            & PPO                & 25.434                  & 1.201                  & 24.521                  & 12.302                  & 1.371                  & 9.203                   & 3.452                   & 2.600                  & 8.021                   & 17.052                                         & 6.158                                           \\
\multicolumn{1}{l|}{\multirow{-2}{*}{LLM + RL}}                                                  & GRPO               & 27.643                  & 0.500                  & 27.421                  & 14.019                  & 0.676                  & 11.025                  & 4.096                   & 2.400                  & 9.287                   & 18.521                 & 6.917                   \\ \hline
\multicolumn{1}{l|}{}                                                                            & Search-o1          & 23.023                  & 17.920                 & 25.702                  & 24.231                  & 17.802                 & 25.680                  & 8.012                   & 5.673                  & 15.982                  & 22.215                                         & 16.230                                          \\ \hhline{~|------------}
\multicolumn{1}{l|}{}                                                                            & R1-Searcher(PPO)     & 37.493                  & 35.982                 & 47.232                  & 34.544                  & 29.982                 & 36.403                  & 9.082                   & 7.230                  & 16.541                  & 40.236                                         & 22.297                                          \\
\multicolumn{1}{l|}{\multirow{-3}{*}{\begin{tabular}[c]{@{}l@{}}LLM + RAG\\ (+RL)\end{tabular}}} & R1-Searcher(GRPO)    & 36.920                  & 32.673                 & 45.392                  & 35.012                  & 30.861                 & 37.620                  & 10.231                  & 7.975                  & 19.200                  & 38.328                                         & 23.483                                          \\ \hline
\multicolumn{1}{l|}{}                                                                            & Llama3.1-8B        & 20.800                  & 14.000                 & 21.452                  & 22.281                  & 15.872                 & 22.427                  & 4.220                   & 2.524                  & 8.044                   & 18.751                                         & 12.561                                          \\
\multicolumn{1}{l|}{\multirow{-2}{*}{LLM + MAS}}                                                 & Qwen2.5-72B        & 32.519                  & 24.254                 & 34.631                  & 36.379                  & 18.511                 & 25.630                  & 7.075                   & 4.717                  & 11.959                  & 30.468                                         & 17.379                                          \\ \hline
\multicolumn{1}{l|}{}                                                                            & MAPPO              & 38.217                  & 34.450                 & 46.400                  & 35.790                  & 31.767                 & 38.722                  & 11.626                  & 9.350                  & 19.738                  & 39.689                                         & 24.499                                          \\ \hhline{~|------------}
\multicolumn{1}{l|}{}                                                                            & \gls{ours}-IS      & 37.677                  & 33.707                 & 45.582                  & 34.796                  & 31.115                 & 38.050                  & 9.888                   & 7.944                  & 18.421                  & 38.989                                         & 23.369                                          \\
\multicolumn{1}{l|}{}                                                                            & \gls{ours}-FoF     & {\ul 40.459}            & {\ul 36.570}           & {\ul 49.429}            & {\ul 36.299}            & {\ul 31.750}           & {\ul 39.089}            & \textbf{12.702}         & \textbf{10.012}        & \textbf{21.633}         & {\ul 42.153}                                   & \textbf{25.248}                                 \\
\multicolumn{1}{l|}{}                                                                            & \gls{ours}-RR      & \textbf{40.864}         & \textbf{37.043}        & \textbf{49.724}         & \textbf{36.840}         & \textbf{31.838}        & \textbf{39.388}         & {\ul 11.378}            & {\ul 9.350}            & {\ul 21.144}            & \textbf{42.544}                                & {\ul 24.990}                                    \\ \hhline{~|------------}
\multicolumn{1}{l|}{\multirow{-5}{*}{\begin{tabular}[c]{@{}l@{}}LLM +MAS \\ +MARL\end{tabular}}} & \gls{ours}-FoF(os) & 41.512                  & 37.623                 & 50.858                  & 37.222                  & 32.801                 & 40.368                  & 12.660                  & 10.923                 & 22.677                  & 43.331                                         & 26.109                                          \\ \hline
\end{tabular}%
}
\end{table*}

\begin{figure}
    \centering
    \includegraphics[width=\linewidth]{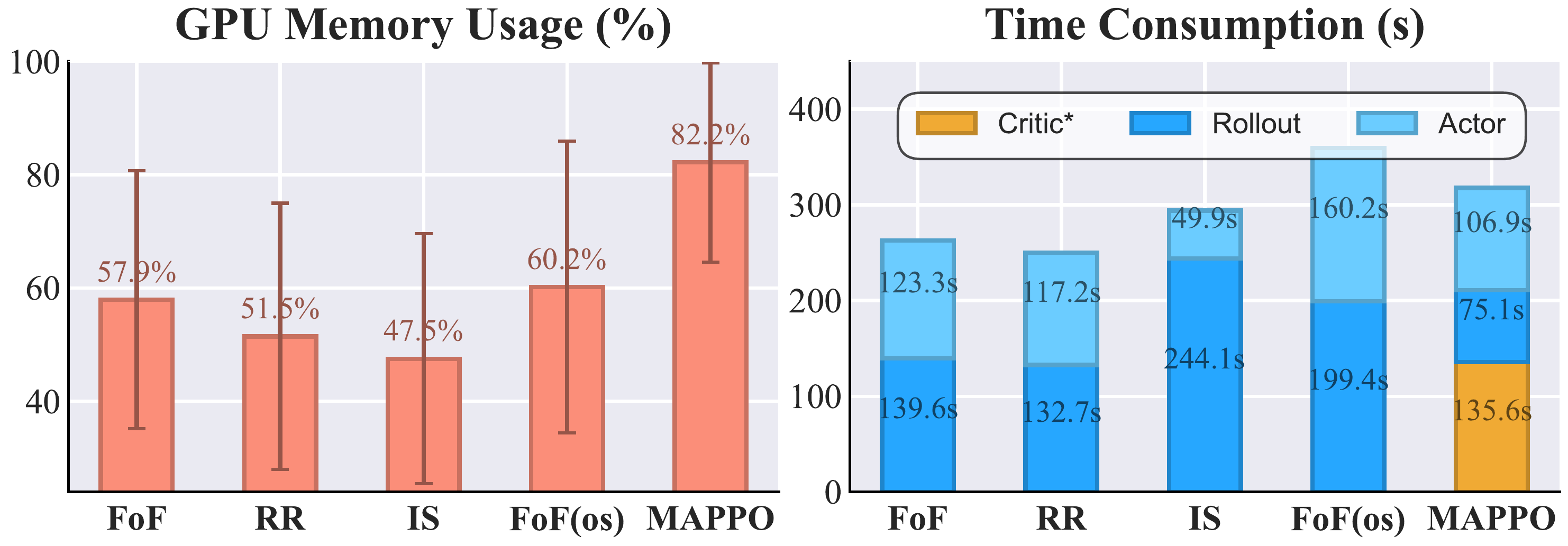}
    \caption{GPU memory usage (\%) and average training-step time (s) for different \gls{marl} algorithms, averaged over the first 50 steps.}
    \label{fig:gpu-timing}
\end{figure}

We evaluate \gls{ours} on HotpotQA with a three-agent \gls{mass} trained with \gls{rl} only, without task-specific \gls{sft} warm-up. We use a batch size of 512 and train for one epoch (176 steps). Baselines include \glspl{mas} instantiated with different \glspl{llm}, an \gls{mappo}-optimized \gls{mas}, and single-context \gls{rag} systems: Search-o1~\citep{DBLP:journals/corr/abs-2501-05366}, which injects search into long-horizon \gls{llm} reasoning, and R1-Searcher~\citep{DBLP:journals/corr/abs-2503-05592}, which further optimizes this pipeline end-to-end with \gls{rl}.


\begin{tcolorbox}[colback=gray!20!white,colframe=black,left=2mm, right=2mm, top=1mm, bottom=1mm, boxsep=0pt]
\textbf{Question 1.} \textit{How does \gls{ours} improve training stability and efficiency compared to \gls{ppo}-style methods?}
\end{tcolorbox}

We visualize learning dynamics by tracking training reward and validation F1 on a fixed 500-example HotpotQA subset over one epoch (Figure~\ref{fig:training}), and quantify training cost separately (Figure~\ref{fig:gpu-timing}). \gls{mappo} exhibits noisier optimization and a lower performance ceiling, while incurring substantially higher memory and wall-clock overhead from maintaining and updating a full critic. In contrast, \gls{ours} converges more smoothly under markedly smaller resource budgets. The \gls{is} variant underperforms because independent optimization within homogeneous groups fails to capture inter-agent dependencies, limiting system-level improvement. By estimating advantages over \emph{heterogeneous} groups, \gls{fof} and \gls{rr} enable end-to-end optimization with global rewards, yielding stronger and more stable convergence with minimal overhead. Notably, \gls{rr} further reduces compute and memory by trading off sample efficiency, yet matches or surpasses the final performance of stronger baselines, making it attractive in large-sample regimes. Finally, \gls{fof}(os) achieves the best validation-F1 trajectory: although slightly weaker in training reward than \gls{fof}, it improves generalization by training longer without increasing the memory footprint. Overall, these results highlight the strong performance--efficiency trade-off and practical scalability of the \gls{ours} family.

\begin{figure}
    \centering
    \includegraphics[width=\linewidth]{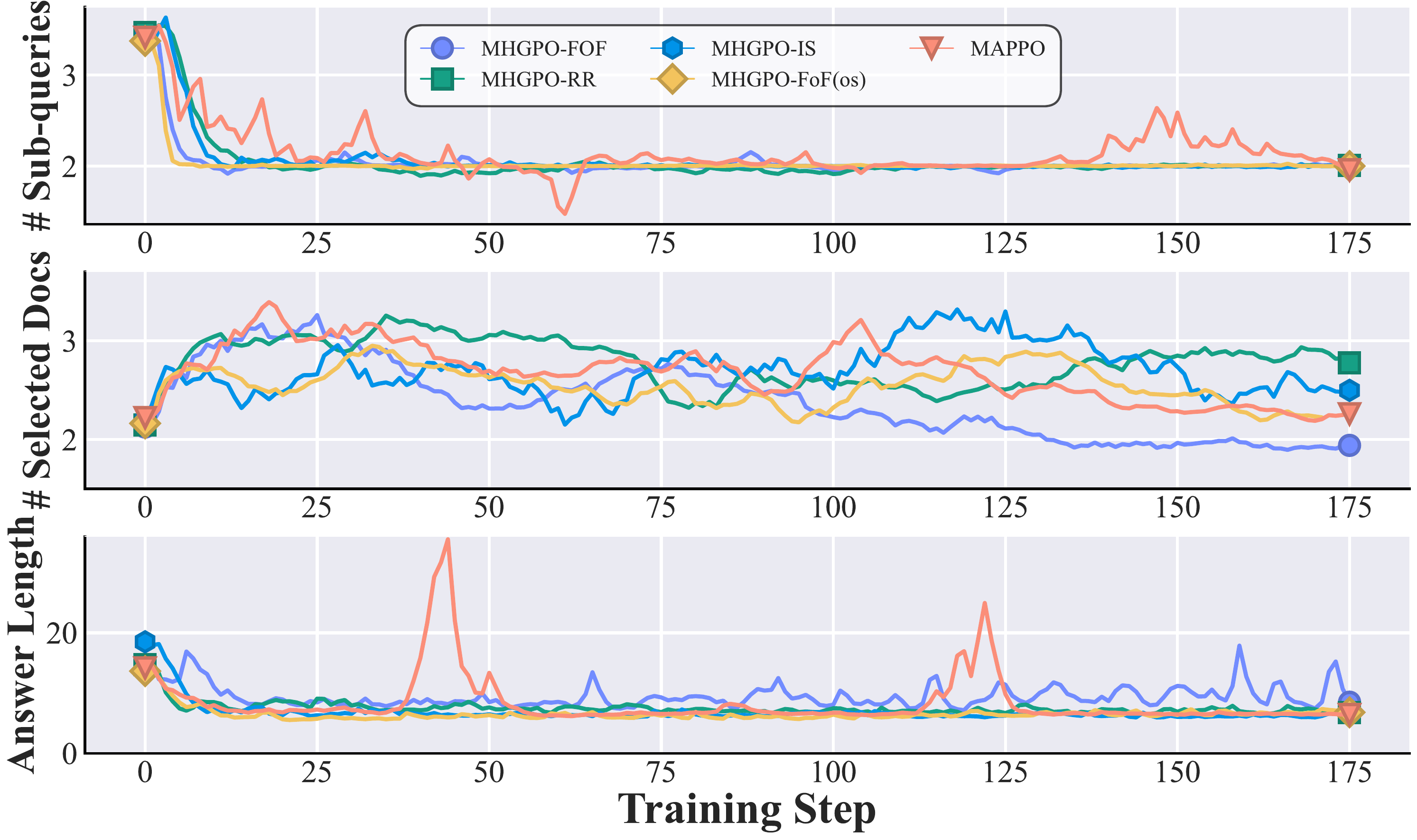}
    \caption{Training dynamics of agent outputs under different \gls{marl} algorithms. Rows correspond to the rewriter, reranker, and answerer (top to bottom).}
    \label{fig:training-agent}
\end{figure}

We also track each agent’s outputs throughout training (Figure~\ref{fig:training-agent}) to show how end-to-end optimization shapes individual behaviors. \gls{mappo} exhibits pronounced instability, plausibly because a single critic must fit diverse objectives across three agents. In contrast, variants of \gls{ours} converge smoothly with higher cross-agent stability: the rewriter consistently produces two rewritten queries, the reranker settles on selecting 2--3 documents, and the final response becomes succinct, with answer length dropping below 10.

\begin{table}[]
\caption{Ablation of role-wise training in \gls{ours} (\gls{fof}) on HotpotQA. We compare full collaborative optimization with training a single role only (\texttt{*}) and freezing one role (\texttt{-}). Results are averaged over three random seeds.}
\label{tab:abla1}
\resizebox{\columnwidth}{!}{%
\begin{tabular}{l|lll
>{\columncolor[HTML]{EFEFEF}}l }
\hline
                     & ACC              & EM               & F1               & AVG                            \\ \hline
\gls{ours} & 40.421{\scriptsize$\pm$0.230} & 36.120{\scriptsize$\pm$0.273} & 49.534{\scriptsize$\pm$0.199} & \textbf{42.025}                \\ \hline
*Rewriter            & 36.798{\scriptsize$\pm$0.560} & 30.481{\scriptsize$\pm$0.450} & 46.151{\scriptsize$\pm$0.492} & 37.810                         \\
*Reranker            & 24.240{\scriptsize$\pm$0.621} & 19.082{\scriptsize$\pm$0.490} & 30.760{\scriptsize$\pm$0.392} & 24.694                         \\
*Answerer            & 21.180{\scriptsize$\pm$0.390} & 18.667{\scriptsize$\pm$0.383} & 29.027{\scriptsize$\pm$0.410} & 22.958                         \\ \hline
-Rewriter            & 31.127{\scriptsize$\pm$0.490} & 23.983{\scriptsize$\pm$0.530} & 33.689{\scriptsize$\pm$0.502} & \cellcolor[HTML]{EFEFEF}29.599 \\
-Reranker            & 38.242{\scriptsize$\pm$0.217} & 34.180{\scriptsize$\pm$0.317} & 46.792{\scriptsize$\pm$0.252} & \cellcolor[HTML]{EFEFEF}39.738 \\
-Answerer            & 38.608{\scriptsize$\pm$0.473} & 35.983{\scriptsize$\pm$0.359} & 48.982{\scriptsize$\pm$0.563} & \cellcolor[HTML]{EFEFEF}41.190 \\ \hline
\end{tabular}%
}
\end{table}

\begin{tcolorbox}[colback=gray!20!white,colframe=black,left=2mm, right=2mm, top=1mm, bottom=1mm, boxsep=0pt]
\textbf{Question 2.} \textit{How does \gls{ours} perform on the test set compared to the baselines? Can \gls{ours} generalize to other datasets?}
\end{tcolorbox}


After \gls{marl} training, we deploy the learned policy as the backbone of our \gls{mas} and evaluate retrieval-intensive QA benchmarks. Table~\ref{tab:main} compares \glspl{mas} optimized by different \gls{rl} algorithms against strong baselines, including standalone \glspl{llm}, unoptimized \glspl{mas}, and single-context \gls{rag} systems; we also report \gls{rl}-tuned \glspl{llm} trained on HotpotQA for direct QA.

Table~\ref{tab:main} shows that standalone \glspl{llm} are weak on these tasks, and na\"ively wrapping them into an \gls{mas} without optimization can further hurt performance due to instruction-following failures, particularly for smaller models. For the Llama3.1-8B backbone, end-to-end \gls{marl} more than doubles the HotpotQA F1 of the corresponding unoptimized \gls{mas} (from 21.45 to 50.86 for \gls{ours}-\gls{fof}(os)). This gap suggests that prompt engineering alone does not capture the co-adaptation among agents or their adaptation to the specific retrieval environment (see the case study in the Appendix for a concrete illustration).

Within \gls{marl}, the \gls{ours} variants substantially outperform \gls{mappo} on the in-domain HotpotQA test set. \gls{fof}(os) achieves the strongest task performance, whereas \gls{rr} offers the best efficiency and remains highly competitive on the main task metrics. On out-of-domain 2WikiMultihopQA and MuSiQue, all \gls{marl} methods improve over the unoptimized baseline in the reported runs, with the \gls{ours} family delivering the most consistent gains across datasets. Moreover, \gls{ours} also outperforms the single-context \gls{rag} baselines and does not exhibit the same degree of \gls{grpo} instability reported by R1-Searcher~\citep{DBLP:journals/corr/abs-2503-05592}, which we attribute to shorter per-context trajectories and implicit (rather than explicit) cross-stage coupling, leading to faster and more reliable group-based convergence.

\subsection{Why Heterogeneous Grouping Works}


\paragraph{Single-Context Transcript Equivalence as a Theoretical Anchor}
Under parameter sharing and reward broadcasting, \gls{ours} sits on a continuum anchored by single-context \gls{grpo}: at one end, an idealized context-sufficiency assumption (A3, Appendix~\ref{app:single_context_equiv}) makes \gls{ours} reduce exactly to \gls{grpo} on a tool-augmented transcript; at the other, realistic \gls{mas} pipelines violate A3 through summarization, filtering, and asymmetric tool outputs, breaking local contexts apart. \textbf{\gls{ours} is designed to operate precisely in this relaxed regime}: rather than enforcing local sufficiency, heterogeneous grouping exploits the cross-context dependencies lost under A3---rollouts sharing the same system input induce trajectory-level couplings that group-wise advantages leverage for global coordination.

Operating in this relaxed regime also introduces additional reward variance: a downstream agent conditioned on a poor upstream prefix may receive low rewards regardless of its own action. Rather than a defect, this is a direct consequence of optimizing system-level coordination over prefix-conditioned local optimality: \textbf{it may implicitly down-weight trajectories where coordination already fails}, thereby steering learning toward globally successful interaction patterns. The \gls{rr} strategy can be interpreted as a bridge in this process by stochastically mixing heterogeneous updates with homogeneous comparisons, giving each agent both global coordination signals and cleaner input-conditioned supervision.

A complementary factor is \emph{progressive homogenization}: as the shared backbone \gls{llm} converges during training, intra-group diversity naturally decreases (Figure~\ref{fig:diversity}, Appendix), so heterogeneous groups increasingly approximate homogeneous ones. This trend is consistent with task-specific convergence, although the current evidence does not by itself rule out entropy collapse. At a minimum, the joint increase in intra-group similarity and validation F1 indicates that homogenization is not immediately accompanied by degraded system performance~\citep{DBLP:journals/corr/abs-2505-22617}.


\subsection{Ablation Study}

\begin{tcolorbox}[colback=gray!20!white,colframe=black,left=2mm,right=2mm,top=1mm,bottom=1mm,boxsep=0pt]
\textbf{Question 3.} \textit{Does \gls{ours} genuinely capture \emph{global} inter-agent interactions, rather than merely improving individual agents?}
\end{tcolorbox}

To directly probe inter-agent interactions, we ablate which agents are optimized (Table~\ref{tab:abla1}). Training a single agent causes sharp drops, and freezing any one agent degrades the system to varying degrees; optimizing all three agents independently reduces exactly to \gls{ours}-\gls{is}, which is already the weakest variant in Table~\ref{tab:main}. Together with the analysis above, this indicates that the gains of \gls{ours} come from \emph{jointly} shaping the three policies under a shared reward, rather than from improving any single agent in isolation, confirming that \gls{ours} captures global inter-agent interactions.

\begin{table}[]
\caption{Ablation on the reward-aggregation operator $\mathrm{Aggr}(\cdot)$ in backward propagation (Eq.~\ref{eq:aggr}) for the branched variants \gls{ours}-\gls{rr} and \gls{ours}-\gls{fof}(os) on HotpotQA.}
\label{tab:abla2}
\resizebox{\columnwidth}{!}{%
\begin{tabular}{l|cc|cc}
\hline
\multicolumn{1}{c|}{\multirow{2}{*}{\begin{tabular}[c]{@{}c@{}}Aggr\\ Function\end{tabular}}} & \multicolumn{2}{c|}{\gls{ours}-\gls{rr}}              & \multicolumn{2}{c}{\gls{ours}-\gls{fof}(os)}           \\ \cline{2-5} 
\multicolumn{1}{c|}{}                                                                         & EM                        & F1                        & EM                         & F1                        \\ \hline
AVG()                                                                                         & \textbf{37.043{\scriptsize$\pm$0.251}} & \textbf{49.724{\scriptsize$\pm$0.408}} & \textbf{37.623{\scriptsize$\pm$0.190}} & \textbf{50.858{\scriptsize$\pm$0.360}} \\ \hline
MAX()                                                                                         & 34.129{\scriptsize$\pm$0.377}          & 46.343{\scriptsize$\pm$0.532}          & 36.704{\scriptsize$\pm$0.368}          & {\ul 49.020{\scriptsize$\pm$0.489}}    \\ \hline
MIN()                                                                                         & {\ul 35.840{\scriptsize$\pm$0.280}}    & {\ul 47.930{\scriptsize$\pm$0.398}}    & {\ul 36.920{\scriptsize$\pm$0.310}}    & 49.842{\scriptsize$\pm$0.452}          \\ \hline
Rand()                                                                                        & 35.643{\scriptsize$\pm$0.730}          & 46.824{\scriptsize$\pm$0.808}          & 36.602{\scriptsize$\pm$0.520}          & 49.013{\scriptsize$\pm$0.674}          \\ \hline
\end{tabular}%
}
\end{table}

On the reward side, we first ablate the aggregation operator $\mathrm{Aggr}(\cdot)$ used in backward reward propagation (Eq.~\ref{eq:aggr}) for the two branched variants, \gls{ours}-\gls{rr} and \gls{ours}-\gls{fof}(os). As shown in Table~\ref{tab:abla2}, \textsc{AVG} yields the strongest EM/F1 across both variants, while \textsc{Max}/\textsc{Min} introduce biased signals and \textsc{Rand} inflates variance; we therefore use \textsc{AVG} as the default. To further stress-test the reward design, we then strip all agent-specific format penalties and retain only the terminal F1. This sparser regime hurts every method but sharpens the contrasts (Table~\ref{tab:abla3}): \gls{ours}-\gls{rr} still holds a slight edge over \gls{fof} (invalid selection 2.8\% vs.\ 3.8\%, F1 45.1\% vs.\ 44.0\%), consistent with its design of mixing heterogeneous groups (for global trajectory optimization) with more homogeneous comparisons (for local stability); in contrast, \gls{mappo} degrades sharply---F1 drops to 35.0\% and the Rewriter hallucinates $>$10 sub-queries with a 5.3\% invalid selection rate---consistent with the view that its critic struggles with credit assignment under sparse, delayed rewards, whereas our group-relative advantage estimation remains more robust.

\begin{table}[]
\caption{Composite reward vs.\ final-reward-only on HotpotQA. ``Final only'' strips all agent-specific format penalties and uses only the terminal F1; for \gls{rr}, Avg aggregation is applied across branches.}
\label{tab:abla3}
\resizebox{\columnwidth}{!}{%
\begin{tabular}{ll|cc|ccc}
\hline
\multirow{2}{*}{Method} & \multirow{2}{*}{Reward} & \multicolumn{2}{c|}{Task} & \multicolumn{3}{c}{Agent Behavior} \\ \cline{3-7}
 &  & EM & F1 & \#\,Sub-Q & \#\,Docs & Inv.\,(\%) \\ \hline
\multirow{2}{*}{\gls{ours}-\gls{fof}} & Composite & \textbf{36.12} & \textbf{49.53} & 2.02 & 1.99 & 0.05 \\
 & Final only & 27.07 & 44.02 & 6.41 & 3.67 & 3.79 \\ \hline
\multirow{2}{*}{\gls{ours}-\gls{rr}} & Composite & \textbf{37.04} & \textbf{49.72} & 2.01 & 2.89 & 0.01 \\
 & Final only & 27.95 & 45.14 & 6.07 & 3.25 & 2.80 \\ \hline
\multirow{2}{*}{\gls{mappo}} & Composite & 34.45 & 46.40 & 2.10 & 2.29 & 0.09 \\
 & Final only & 18.38 & 35.00 & 10.34 & 3.91 & 5.33 \\ \hline
\end{tabular}%
}
\end{table}




\section{Related Work}

\paragraph{\gls{llm}-based Multi-agent Systems}
Beyond a standalone \gls{llm}, \glspl{mas}---teams of interacting agents, potentially powered by different \glspl{llm}---can better decompose complex objectives and coordinate specialized reasoning. Emerging agent-development frameworks such as AutoGen~\citep{autogen} and OWL~\citep{owl} further highlight the growing practicality of \glspl{mas} in real-world deployments, including financial decision-making and market-game simulation~\citep{DBLP:conf/emnlp/ChenZWWSCX25, DBLP:journals/corr/abs-2602-00948}. In search-based QA, prior work consistently reports gains from collaboration: \citet{DBLP:journals/corr/abs-2501-00332} cast \gls{rag} as a sequential \gls{mas}, using structured inter-agent interaction to improve retrieval-augmented generation; \citet{DBLP:journals/corr/abs-2501-07813} introduce dynamic routing that assigns subtasks to expert agents; and \citet{DBLP:journals/corr/abs-2503-06951} incorporate backtracking to support reversible multi-hop reasoning, strengthening \gls{rag}-based QA.

\paragraph{End-to-end \gls{rl} for \glspl{mas}}
Despite their promise, \glspl{mas} remain brittle in practice: \citet{DBLP:journals/corr/abs-2503-13657} attribute failures to fragile system design and inconsistent \gls{llm} formatting. This motivates end-to-end reinforcement learning for \glspl{mas} (i.e., \gls{marl}) to learn robust, adaptive coordination instead of hand-crafted control logic. The dominant paradigm, centralized training with decentralized execution, is represented by \gls{mappo}~\citep{DBLP:conf/nips/YuVVGWBW22} and HAPPO~\citep{DBLP:conf/iclr/KubaCWWSW022}, which often adopt \emph{parameter sharing} for scalable training. In \gls{llm}-based \glspl{mas}, recent RL efforts include component-level optimization~\citep{DBLP:journals/corr/abs-2401-06800,DBLP:journals/corr/abs-2305-14283}, applying \gls{mappo} to multi-agent \gls{rag} with a shared \gls{llm} backbone~\citep{DBLP:journals/corr/abs-2501-15228}, and tailoring RL frameworks for end-to-end optimization across interacting agents~\citep{DBLP:journals/corr/abs-2504-16129}.

\paragraph{Critic-free Group-based Policy Optimization}
Recent critic-free \gls{rl} methods such as \gls{grpo} replace explicit value estimation with reward normalization over sampled groups, substantially reducing optimization overhead in \gls{llm} training~\citep{DBLP:journals/corr/abs-2402-03300, DBLP:journals/corr/abs-2503-14476}. These methods are typically developed for \emph{single-context} reasoning, where all responses in a group are conditioned on the same prompt and are compared under matched local contexts. Our work differs in both setting and objective: we study how group-relative optimization can be extended to \emph{multi-agent, multi-context} systems, where rollouts across agents no longer share identical local prompts and the goal is to optimize global system success rather than per-agent responses under a fixed prefix.
\section{Conclusions}

In this paper, we propose \gls{ours}, a critic-free reinforcement learning framework for end-to-end optimization of multi-agent \gls{llm} systems. The core idea is to combine backward reward propagation, which exposes indirect inter-agent dependencies, with heterogeneous-group advantage estimation, which shifts optimization from local agent behavior toward global system success. We further show how the framework relates to single-context \gls{grpo} under an idealized context-sufficiency assumption, while being designed for practical search settings with incomplete local contexts. Experiments on multi-hop QA benchmarks demonstrate that the \gls{ours} family achieves a strong performance--efficiency trade-off over strong baselines, with different variants favoring task performance or training efficiency. Although our evaluation focuses on \gls{mass}, the framework is readily extensible to broader \glspl{mas}; future work will test this scalability in more diverse tasks and larger agent teams.

\section{Limitations}
\paragraph{Task scope (QA-only).} Our evaluation is restricted to multi-hop QA with static iterative retrieval, and does not cover open-ended agentic search, long-horizon planning, or tool-use-heavy settings. The generality of MHGPO beyond QA-centric pipelines---particularly under dynamic, non-stationary environments---remains to be explored.

\paragraph{Theoretical characterization.} We provide a formal bridge to single-context group-based optimization under an idealized context-sufficiency assumption, but leave a full variance and stability analysis of heterogeneous grouping under incomplete local contexts to future work.

\paragraph{Model scale (up to 8B).} Due to computational constraints, all experiments use backbones up to 8B parameters. Scaling behavior and emergent coordination effects with larger \glspl{llm} remain untested.

\paragraph{Fixed agent count (3 agents).} All experiments use a three-agent pipeline (Rewriter $\rightarrow$ Reranker $\rightarrow$ Answerer) to match standard RAG-QA baselines. Although MHGPO is formulated for arbitrary $n$, systematic evaluation under varying team sizes and non-linear topologies (e.g., branching or graph-structured coordination) is left to future work.

\bibliography{resources/references}

\begin{thebibliography}{28}
\providecommand{\natexlab}[1]{#1}

\bibitem[{Cemri et~al.(2025)Cemri, Pan, Yang, Agrawal, Chopra, Tiwari, Keutzer,
  Parameswaran, Klein, Ramchandran, Zaharia, Gonzalez, and
  Stoica}]{DBLP:journals/corr/abs-2503-13657}
Mert Cemri, Melissa~Z. Pan, Shuyi Yang, Lakshya~A. Agrawal, Bhavya Chopra,
  Rishabh Tiwari, Kurt Keutzer, Aditya~G. Parameswaran, Dan Klein, Kannan
  Ramchandran, Matei Zaharia, Joseph~E. Gonzalez, and Ion Stoica. 2025.
\newblock \href {https://doi.org/10.48550/ARXIV.2503.13657} {Why do multi-agent
  {LLM} systems fail?}
\newblock \emph{CoRR}, abs/2503.13657.

\bibitem[{Chang et~al.(2025)Chang, Jiang, Rakesh, Pan, Yeh, Wang, Hu, Xu,
  Zheng, Das, and Zou}]{DBLP:journals/corr/abs-2501-00332}
Chia{-}Yuan Chang, Zhimeng Jiang, Vineeth Rakesh, Menghai Pan,
  Chin{-}Chia~Michael Yeh, Guanchu Wang, Mingzhi Hu, Zhichao Xu, Yan Zheng,
  Mahashweta Das, and Na~Zou. 2025.
\newblock \href {https://doi.org/10.48550/ARXIV.2501.00332} {{MAIN-RAG:}
  multi-agent filtering retrieval-augmented generation}.
\newblock \emph{CoRR}, abs/2501.00332.

\bibitem[{Chen et~al.(2025{\natexlab{a}})Chen, Zou, Wang, Wang, Sun, Chi, and
  Xu}]{DBLP:conf/emnlp/ChenZWWSCX25}
Jiaxiang Chen, Mingxi Zou, Zhuo Wang, Qifan Wang, Danny~Dongning Sun, Zhang
  Chi, and Zenglin Xu. 2025{\natexlab{a}}.
\newblock \href {https://aclanthology.org/2025.findings-emnlp.87/} {Finhear:
  Human expertise and adaptive risk-aware temporal reasoning for financial
  decision-making}.
\newblock In \emph{Findings of the Association for Computational Linguistics:
  {EMNLP} 2025, Suzhou, China, November 4-9, 2025}, pages 1648--1672.
  Association for Computational Linguistics.

\bibitem[{Chen et~al.(2025{\natexlab{b}})Chen, Yan, Sun, Ma, Zhang, Wang, Yin,
  Yang, and Mao}]{DBLP:journals/corr/abs-2501-15228}
Yiqun Chen, Lingyong Yan, Weiwei Sun, Xinyu Ma, Yi~Zhang, Shuaiqiang Wang,
  Dawei Yin, Yiming Yang, and Jiaxin Mao. 2025{\natexlab{b}}.
\newblock \href {https://doi.org/10.48550/ARXIV.2501.15228} {Improving
  retrieval-augmented generation through multi-agent reinforcement learning}.
\newblock \emph{CoRR}, abs/2501.15228.

\bibitem[{Cui et~al.(2025)Cui, Zhang, Chen, Yuan, Wang, Zuo, Li, Fan, Chen,
  Chen, Liu, Peng, Bai, Ouyang, Cheng, Zhou, and
  Ding}]{DBLP:journals/corr/abs-2505-22617}
Ganqu Cui, Yuchen Zhang, Jiacheng Chen, Lifan Yuan, Zhi Wang, Yuxin Zuo,
  Haozhan Li, Yuchen Fan, Huayu Chen, Weize Chen, Zhiyuan Liu, Hao Peng, Lei
  Bai, Wanli Ouyang, Yu~Cheng, Bowen Zhou, and Ning Ding. 2025.
\newblock \href {https://doi.org/10.48550/ARXIV.2505.22617} {The entropy
  mechanism of reinforcement learning for reasoning language models}.
\newblock \emph{CoRR}, abs/2505.22617.

\bibitem[{Gao et~al.(2024)Gao, Xiong, Gao, Jia, Pan, Bi, Dai, Sun, Wang, and
  Wang}]{rag-survey}
Yunfan Gao, Yun Xiong, Xinyu Gao, Kangxiang Jia, Jinliu Pan, Yuxi Bi, Yi~Dai,
  Jiawei Sun, Meng Wang, and Haofen Wang. 2024.
\newblock \href {https://arxiv.org/abs/2312.10997} {Retrieval-augmented
  generation for large language models: A survey}.
\newblock \emph{Preprint}, arXiv:2312.10997.

\bibitem[{Guo et~al.(2025)Guo, Yang, Zhang, Song, Wang, Zhu, Xu, Zhang, Ma, Bi,
  Zhang, Yu, Wu, Wu, Gou, Shao, Li, Gao
  et~al.}]{DBLP:journals/nature/GuoYZSWZXZMBZY025}
Daya Guo, Dejian Yang, Haowei Zhang, Junxiao Song, Peiyi Wang, Qihao Zhu,
  Runxin Xu, Ruoyu Zhang, Shirong Ma, Xiao Bi, Xiaokang Zhang, Xingkai Yu,
  Yu~Wu, Z.~F. Wu, Zhibin Gou, Zhihong Shao, Zhuoshu Li, Ziyi Gao, and 1
  others. 2025.
\newblock \href {https://doi.org/10.1038/S41586-025-09422-Z} {Deepseek-r1
  incentivizes reasoning in {LLMs} through reinforcement learning}.
\newblock \emph{Nat.}, 645(8081):633--638.

\bibitem[{Ho et~al.(2020)Ho, Nguyen, Sugawara, and Aizawa}]{2wiki}
Xanh Ho, Anh{-}Khoa~Duong Nguyen, Saku Sugawara, and Akiko Aizawa. 2020.
\newblock \href {https://doi.org/10.18653/V1/2020.COLING-MAIN.580}
  {Constructing {A} multi-hop {QA} dataset for comprehensive evaluation of
  reasoning steps}.
\newblock In \emph{Proceedings of the 28th International Conference on
  Computational Linguistics, {COLING} 2020, Barcelona, Spain (Online), December
  8-13, 2020}, pages 6609--6625. International Committee on Computational
  Linguistics.

\bibitem[{Hu et~al.(2025)Hu, Zhou, Wendong~Fan, Xia, Sun, Ye, Jin, Li, Zhang,
  Wang, Ye, Luo, and Li}]{owl}
Mengkang Hu, Yuhang Zhou, Yuzhou~Nie Wendong~Fan, Bowei Xia, Tao Sun, Ziyu Ye,
  Zhaoxuan Jin, Yingru Li, Zeyu Zhang, Yifeng Wang, Qianshuo Ye, Ping Luo, and
  Guohao Li. 2025.
\newblock \href {https://github.com/camel-ai/owl} {Owl: Optimized workforce
  learning for general multi-agent assistance in real-world task automation}.

\bibitem[{Izacard et~al.(2022)Izacard, Caron, Hosseini, Riedel, Bojanowski,
  Joulin, and Grave}]{contriever}
Gautier Izacard, Mathilde Caron, Lucas Hosseini, Sebastian Riedel, Piotr
  Bojanowski, Armand Joulin, and Edouard Grave. 2022.
\newblock \href {https://openreview.net/forum?id=jKN1pXi7b0} {Unsupervised
  dense information retrieval with contrastive learning}.
\newblock \emph{Trans. Mach. Learn. Res.}, 2022.

\bibitem[{Kuba et~al.(2022)Kuba, Chen, Wen, Wen, Sun, Wang, and
  Yang}]{DBLP:conf/iclr/KubaCWWSW022}
Jakub~Grudzien Kuba, Ruiqing Chen, Muning Wen, Ying Wen, Fanglei Sun, Jun Wang,
  and Yaodong Yang. 2022.
\newblock \href {https://openreview.net/forum?id=EcGGFkNTxdJ} {Trust region
  policy optimisation in multi-agent reinforcement learning}.
\newblock In \emph{The Tenth International Conference on Learning
  Representations, {ICLR} 2022, Virtual Event, April 25-29, 2022}.
  OpenReview.net.

\bibitem[{Kulkarni et~al.(2024)Kulkarni, Tangarajan, Kim, and
  Trivedi}]{DBLP:journals/corr/abs-2401-06800}
Mandar Kulkarni, Praveen Tangarajan, Kyung Kim, and Anusua Trivedi. 2024.
\newblock \href {https://doi.org/10.48550/ARXIV.2401.06800} {Reinforcement
  learning for optimizing {RAG} for domain chatbots}.
\newblock \emph{CoRR}, abs/2401.06800.

\bibitem[{Li et~al.(2025)Li, Dong, Jin, Zhang, Zhou, Zhu, Zhang, and
  Dou}]{DBLP:journals/corr/abs-2501-05366}
Xiaoxi Li, Guanting Dong, Jiajie Jin, Yuyao Zhang, Yujia Zhou, Yutao Zhu,
  Peitian Zhang, and Zhicheng Dou. 2025.
\newblock \href {https://doi.org/10.48550/ARXIV.2501.05366} {Search-o1: Agentic
  search-enhanced large reasoning models}.
\newblock \emph{CoRR}, abs/2501.05366.

\bibitem[{Liao et~al.(2025{\natexlab{a}})Liao, Wen, Wang, and
  Zhang}]{liao2025marftmultiagentreinforcementfinetuning}
Junwei Liao, Muning Wen, Jun Wang, and Weinan Zhang. 2025{\natexlab{a}}.
\newblock \href {https://arxiv.org/abs/2504.16129} {Marft: Multi-agent
  reinforcement fine-tuning}.
\newblock \emph{Preprint}, arXiv:2504.16129.

\bibitem[{Liao et~al.(2025{\natexlab{b}})Liao, Wen, Wang, and
  Zhang}]{DBLP:journals/corr/abs-2504-16129}
Junwei Liao, Muning Wen, Jun Wang, and Weinan Zhang. 2025{\natexlab{b}}.
\newblock \href {https://doi.org/10.48550/ARXIV.2504.16129} {{MARFT:}
  multi-agent reinforcement fine-tuning}.
\newblock \emph{CoRR}, abs/2504.16129.

\bibitem[{Ma et~al.(2023)Ma, Gong, He, Zhao, and
  Duan}]{DBLP:journals/corr/abs-2305-14283}
Xinbei Ma, Yeyun Gong, Pengcheng He, Hai Zhao, and Nan Duan. 2023.
\newblock \href {https://doi.org/10.48550/ARXIV.2305.14283} {Query rewriting
  for retrieval-augmented large language models}.
\newblock \emph{CoRR}, abs/2305.14283.

\bibitem[{Schulman et~al.(2017)Schulman, Wolski, Dhariwal, Radford, and
  Klimov}]{DBLP:journals/corr/SchulmanWDRK17}
John Schulman, Filip Wolski, Prafulla Dhariwal, Alec Radford, and Oleg Klimov.
  2017.
\newblock \href {https://arxiv.org/abs/1707.06347} {Proximal policy
  optimization algorithms}.
\newblock \emph{CoRR}, abs/1707.06347.

\bibitem[{Shao et~al.(2024)Shao, Wang, Zhu, Xu, Song, Zhang, Li, Wu, and
  Guo}]{DBLP:journals/corr/abs-2402-03300}
Zhihong Shao, Peiyi Wang, Qihao Zhu, Runxin Xu, Junxiao Song, Mingchuan Zhang,
  Y.~K. Li, Y.~Wu, and Daya Guo. 2024.
\newblock \href {https://doi.org/10.48550/ARXIV.2402.03300} {Deepseekmath:
  Pushing the limits of mathematical reasoning in open language models}.
\newblock \emph{CoRR}, abs/2402.03300.

\bibitem[{Song et~al.(2025)Song, Jiang, Min, Chen, Chen, Zhao, Fang, and
  Wen}]{DBLP:journals/corr/abs-2503-05592}
Huatong Song, Jinhao Jiang, Yingqian Min, Jie Chen, Zhipeng Chen, Wayne~Xin
  Zhao, Lei Fang, and Ji{-}Rong Wen. 2025.
\newblock \href {https://doi.org/10.48550/ARXIV.2503.05592} {R1-searcher:
  Incentivizing the search capability in llms via reinforcement learning}.
\newblock \emph{CoRR}, abs/2503.05592.

\bibitem[{Su et~al.(2024)Su, Tang, Ai, Wu, and Liu}]{DBLP:conf/acl/SuTA0024}
Weihang Su, Yichen Tang, Qingyao Ai, Zhijing Wu, and Yiqun Liu. 2024.
\newblock \href {https://doi.org/10.18653/V1/2024.ACL-LONG.702} {{DRAGIN:}
  dynamic retrieval augmented generation based on the real-time information
  needs of large language models}.
\newblock In \emph{Proceedings of the 62nd Annual Meeting of the Association
  for Computational Linguistics (Volume 1: Long Papers), {ACL} 2024, Bangkok,
  Thailand, August 11-16, 2024}, pages 12991--13013. Association for
  Computational Linguistics.

\bibitem[{Trivedi et~al.(2022)Trivedi, Balasubramanian, Khot, and
  Sabharwal}]{musique}
Harsh Trivedi, Niranjan Balasubramanian, Tushar Khot, and Ashish Sabharwal.
  2022.
\newblock \href {https://doi.org/10.1162/TACL\_A\_00475} {Musique: Multihop
  questions via single-hop question composition}.
\newblock \emph{Trans. Assoc. Comput. Linguistics}, 10:539--554.

\bibitem[{Wu et~al.(2025)Wu, Li, Wei, Li, Ding, and
  Gao}]{DBLP:journals/corr/abs-2501-07813}
Feijie Wu, Zitao Li, Fei Wei, Yaliang Li, Bolin Ding, and Jing Gao. 2025.
\newblock \href {https://doi.org/10.48550/ARXIV.2501.07813} {Talk to right
  specialists: Routing and planning in multi-agent system for question
  answering}.
\newblock \emph{CoRR}, abs/2501.07813.

\bibitem[{Wu et~al.(2023)Wu, Bansal, Zhang, Wu, Zhang, Zhu, Li, Jiang, Zhang,
  and Wang}]{autogen}
Qingyun Wu, Gagan Bansal, Jieyu Zhang, Yiran Wu, Shaokun Zhang, Erkang Zhu,
  Beibin Li, Li~Jiang, Xiaoyun Zhang, and Chi Wang. 2023.
\newblock \href {https://doi.org/10.48550/ARXIV.2308.08155} {Autogen: Enabling
  next-gen {LLM} applications via multi-agent conversation framework}.
\newblock \emph{CoRR}, abs/2308.08155.

\bibitem[{Yang et~al.(2018)Yang, Qi, Zhang, Bengio, Cohen, Salakhutdinov, and
  Manning}]{hotpot}
Zhilin Yang, Peng Qi, Saizheng Zhang, Yoshua Bengio, William~W. Cohen, Ruslan
  Salakhutdinov, and Christopher~D. Manning. 2018.
\newblock \href {https://doi.org/10.18653/V1/D18-1259} {Hotpotqa: {A} dataset
  for diverse, explainable multi-hop question answering}.
\newblock In \emph{Proceedings of the 2018 Conference on Empirical Methods in
  Natural Language Processing, Brussels, Belgium, October 31 - November 4,
  2018}, pages 2369--2380. Association for Computational Linguistics.

\bibitem[{Yu et~al.(2022)Yu, Velu, Vinitsky, Gao, Wang, Bayen, and
  Wu}]{DBLP:conf/nips/YuVVGWBW22}
Chao Yu, Akash Velu, Eugene Vinitsky, Jiaxuan Gao, Yu~Wang, Alexandre~M. Bayen,
  and Yi~Wu. 2022.
\newblock \href
  {http://papers.nips.cc/paper\_files/paper/2022/hash/9c1535a02f0ce079433344e14d910597-Abstract-Datasets\_and\_Benchmarks.html}
  {The surprising effectiveness of {PPO} in cooperative multi-agent games}.
\newblock In \emph{Advances in Neural Information Processing Systems 35: Annual
  Conference on Neural Information Processing Systems 2022, NeurIPS 2022, New
  Orleans, LA, USA, November 28 - December 9, 2022}.

\bibitem[{Yu et~al.(2025)Yu, Zhang, Zhu, Yuan, Zuo, Yue, Fan, Liu, Liu, Liu,
  Lin, Lin, Ma, Sheng, Tong, Zhang, Zhang, Zhang, Zhu, Zhu, Chen, Chen, Wang,
  Yu, Dai, Song, Wei, Zhou, Liu, Ma, Zhang, Yan, Qiao, Wu, and
  Wang}]{DBLP:journals/corr/abs-2503-14476}
Qiying Yu, Zheng Zhang, Ruofei Zhu, Yufeng Yuan, Xiaochen Zuo, Yu~Yue, Tiantian
  Fan, Gaohong Liu, Lingjun Liu, Xin Liu, Haibin Lin, Zhiqi Lin, Bole Ma,
  Guangming Sheng, Yuxuan Tong, Chi Zhang, Mofan Zhang, Wang Zhang, Hang Zhu,
  and 16 others. 2025.
\newblock \href {https://doi.org/10.48550/ARXIV.2503.14476} {{DAPO:} an
  open-source {LLM} reinforcement learning system at scale}.
\newblock \emph{CoRR}, abs/2503.14476.

\bibitem[{Zhao et~al.(2025)Zhao, Gao, Yang, Chen, Wang, Zhu, Tang, and
  Li}]{DBLP:journals/corr/abs-2503-06951}
Xinjie Zhao, Fan Gao, Rui Yang, Yingjian Chen, Yuyang Wang, Ying Zhu, Jiacheng
  Tang, and Irene Li. 2025.
\newblock \href {https://doi.org/10.48550/ARXIV.2503.06951} {Reagent:
  Reversible multi-agent reasoning for knowledge-enhanced multi-hop {QA}}.
\newblock \emph{CoRR}, abs/2503.06951.

\bibitem[{Zou et~al.(2026)Zou, Chen, Luo, Dai, Zhang, Sun, and
  Xu}]{DBLP:journals/corr/abs-2602-00948}
Mingxi Zou, Jiaxiang Chen, Aotian Luo, Jingyi Dai, Chi Zhang, Dongning Sun, and
  Zenglin Xu. 2026.
\newblock \href {https://doi.org/10.48550/ARXIV.2602.00948} {Finevo: From
  isolated backtests to ecological market games for multi-agent financial
  strategy evolution}.
\newblock \emph{CoRR}, abs/2602.00948.

\end{thebibliography}
\clearpage

\appendix

\section{Detailed Algorithm}

\subsection{Multi-Agent Policy Optimization}

The complete optimization procedure of the proposed \gls{ours} framework is outlined in Algorithm~\ref{algo:ours}. Central to this process is the sampling function $\mathcal{H}(\theta)$, which can be instantiated using one of the \gls{is}, \gls{fof}, or \gls{rr} strategies. This function is pivotal for trajectory sampling within the \gls{mas}, as it determines both the sampling behavior and the group assignment for each rollout. In the prototypical \gls{mass} considered in this work, the reward function $\mathcal{R}$ is instantiated as the F1 score.

\subsection{Rollout Sampling Strategies}
To introduce the proposed rollout sampling algorithms, we begin by presenting a core utility function, \textit{fork\_on()}, defined in Algorithm~\ref{algo:fo}. This function facilitates group rollout by conducting one-to-one trajectory sampling before and after the \emph{first interaction with Agent~$i$}, and performing a one-to-many branching exclusively at Agent~$i$, designated as the fork point. Concretely, the function generates $G$ parallel trajectories starting from the fork point and assigns group identifiers based on the index of the forking agent. The output is a complete set of group-specific forked trajectories.

\subsubsection{\acrfull{fof}}
The \gls{fof} strategy leverages this utility by directly invoking \textit{fork\_on()} with the first agent in the trajectory designated as the fork point, as detailed in Algorithm~\ref{algo:fof}. 


\subsubsection{\acrfull{is}} In contrast, \gls{is} executes an exhaustive search across all agents for each input $q$, invoking \textit{fork\_on()} at every potential fork point. After sampling, it filters out all \emph{heterogeneous} rollout groups and retains only the homogeneous ones for downstream training.

\subsubsection{\acrfull{rr}}
The \gls{rr} method operates at the batch level. For each sample in the batch, it randomly selects an Agent~$i$ as the fork point according to a predefined probability distribution $\{p_i\}$. Upon processing the entire batch, rollout samples that belong to singleton groups (i.e., groups with no shared samples) are re-shuffled and regrouped prior to training, as specified in Algorithm~\ref{algo:rr}.

\section{Implementation Details}

All experiments were run on a single machine with eight NVIDIA H100 GPUs (80\,GB HBM3 each) and an Intel Xeon Platinum 8468 CPU. For the \gls{mappo} baseline, we instantiate the critic with the same backbone \gls{llm} architecture as the actor to ensure comparable modeling capacity. For R1-Searcher and Search-o1, we use their official default implementations and modify only two components: the backbone \gls{llm} and the retrieval engine.

\subsection{Training Dataset}
We utilized the training set of HotpotQA~\citep{hotpot}, comprising 90,447 samples, as the reinforcement learning training data. Only the questions from the dataset were used as model inputs, with the corresponding answers serving as ground-truth labels for computing rewards based on the F1 score. Representative examples are presented in Figure~\ref{tab:exp-hotpot}. These questions require the model to invoke external tools for search and perform multi-hop reasoning based on the retrieved results, effectively evaluating the agent's tool-calling capabilities.

\subsection{Hyper-parameters}
The hyperparameter ranges and detailed configurations used in all experiments are presented in Table~\ref{tab:hps}. Specifically, the critic network for \gls{mappo} employs a learning rate of 1e-6. For the actor networks in both \gls{mappo} and \gls{ours}, learning rates are selected from {1e-6, 5e-7}, with 5e-7 empirically demonstrating superior performance. All other parameters are shared across different \gls{marl} algorithms.
\begin{table}[]
\caption{Key hyperparameters for the implementation of \gls{ours} and MAPPO.}
\label{tab:hps}
\resizebox{\columnwidth}{!}{%
\begin{tabular}{lll}
\hline
\textbf{Hyperparameter} & \textbf{Explanation}       & \textbf{Value(s)} \\ \hline
\texttt{lr}             & learning rate              & (5e-7, 1e-6)           \\
\texttt{rollout.n}      & group size for \glspl{goa} & 4              \\
\texttt{batch\_size}    & batch size                 & 512            \\
$\epsilon$              & clip range                 & 0.2            \\
\texttt{top\_n}         & param for top-n sampling   & 0.9            \\
$\beta$                 & param for KL penalty       & 0.001          \\ \hline
\end{tabular}%
}
\end{table}
\begin{table}[H]
\caption{HotpotQA performance of \gls{ours} with varying group sizes, averaged over three random seeds.}
\label{tab:group}
\resizebox{\columnwidth}{!}{%
\begin{tabular}{l|lll
>{\columncolor[HTML]{EFEFEF}}l }
\hline
Group Size & ACC                       & EM                        & F1                        & \cellcolor[HTML]{EFEFEF}AVG \\ \hline
3          & 36.421$\pm$0.394          & 34.120$\pm$0.320          & 43.534$\pm$0.262          & 38.025                      \\
4          & {\ul 40.421$\pm$0.230}    & {\ul 36.120$\pm$0.273}    & \textbf{49.534$\pm$0.199} & {\ul 42.025}                \\
5          & 39.989$\pm$0.170          & 36.021$\pm$0.189          & 49.379$\pm$0.210          & 41.796                      \\
6          & \textbf{41.030$\pm$0.190} & \textbf{36.400$\pm$0.202} & {\ul 49.422$\pm$0.238}    & \textbf{42.284}             \\ \hline
\end{tabular}%
}
\end{table}
\subsection{Agent Prompts}
We follow the three-agent \gls{mas} design proposed by \citet{DBLP:journals/corr/abs-2501-15228}, with only minor modifications to the agents' prompts and output formatting to simplify the system implementation.

\begin{tcolorbox}[colback=gray!5!white, colframe=black, width=\linewidth,
title=Prompt for Rewriter]
\textbf{System:} You are a professional assistant proficient in transforming complex or unclear questions into simpler, more searchable sub-questions.

Please assist me in rewriting or breaking down the provided questions into sub-questions to make it easier to search for answers using a search engine. The rewritten sub-questions must have logical connections and dependencies, and should not be overly repetitive in meaning.

Output the sub-questions in the form of a string list, with the format: 
\begin{quote}
\texttt{\#\#\# <query1>; <query2>; <query3>; ... \#\#\#}
\end{quote}


\textbf{User:} Original question is \{\texttt{raw\_question...}\}
\end{tcolorbox}
\begin{tcolorbox}[colback=gray!5!white, colframe=black, width=\linewidth, title=Prompt for Reranker]
\textbf{System:} You are a helpful, respectful, and honest assistant. Your task is to identify and output the IDs (such as 0, 1, 2, etc.) of the candidate Documents that are useful for answering the given Question.

Now, simply output the IDs of those candidate Documents that can assist in answering the Question: \{\texttt{raw\_question}\}. Arrange them in the order where the more relevant documents are listed first.

\textbf{User:} Question is: \{\texttt{raw\_question}\}

Document0: \{\texttt{title:..., snippet:...}\}

Document1: \{\texttt{title:..., snippet:...}\}

\ldots

Do not output anything else.
\end{tcolorbox}

\begin{tcolorbox}[colback=gray!5!white, colframe=black, width=\linewidth,
title=Prompt for Answerer]
\textbf{System:} You are a helpful, respectful and honest assistant. Your task is to predict the answer to the question based on the given documents. If you don't know the answer to a question, please don't share false information. Answer the question as accurately as possible.

\textbf{User:} Question is: \{\texttt{raw\_question}\}

Document 0: \{\texttt{title: ..., snippet: ...}\}

Document 1: \{\texttt{title: ..., snippet: ...}\}

\ldots

Now, answer the Question: \{\texttt{raw\_question}\}, based on the above Documents.
\end{tcolorbox}

\begin{figure}
    \centering
\includegraphics[width=1\linewidth]{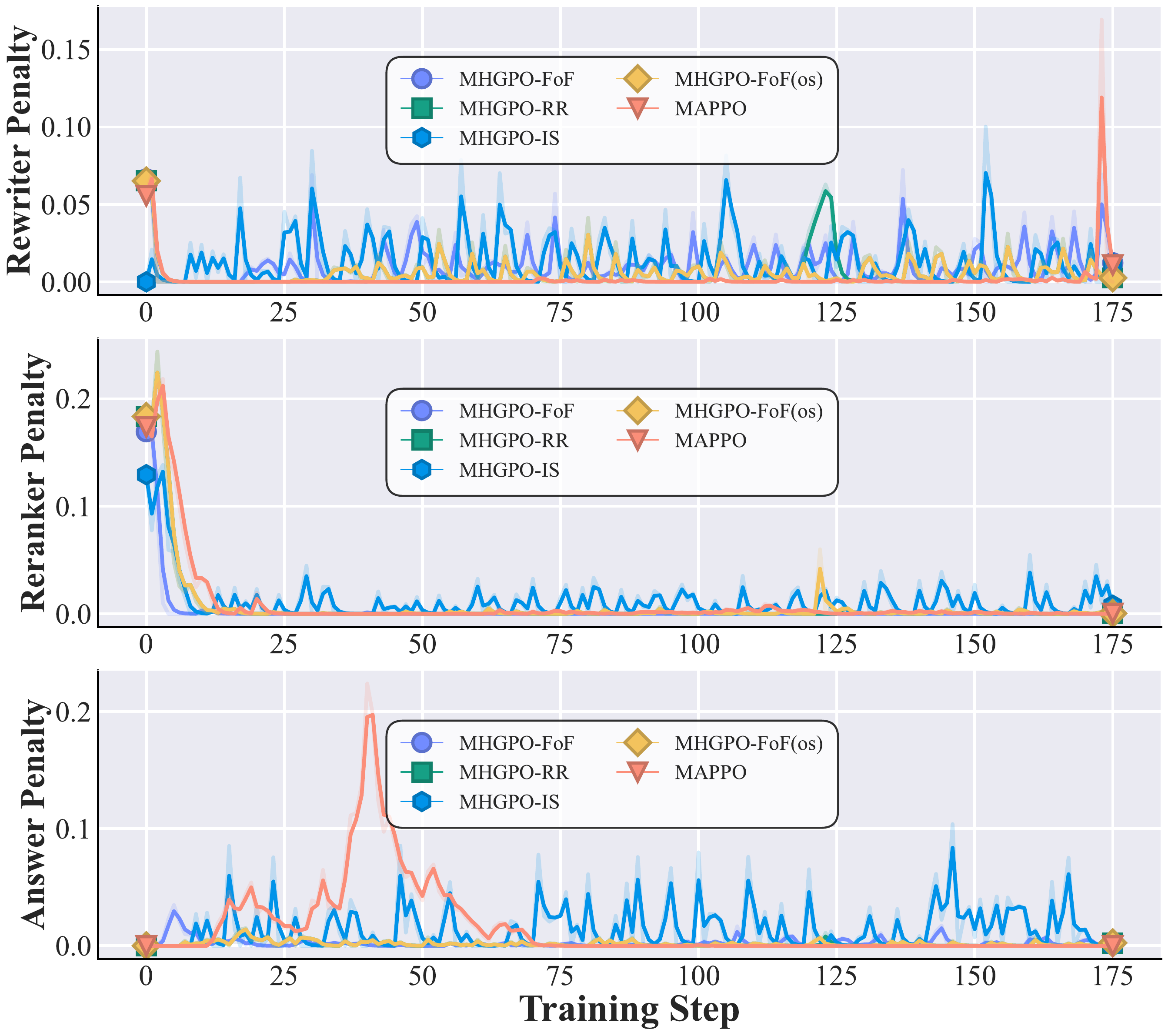}
    \caption{Agent-specific rewards (format penalty) for each agent in the three-agent \gls{mas} during one training epoch on HotpotQA using different \gls{marl} algorithms.}
    \label{fig:penalty}

\end{figure}

\subsection{Agent-specific Rewards}
For agent-specific reward design, we primarily employ a format reward to enforce output standardization for each agent, ensuring seamless functionality across intermediate system interfaces. Specifically, the Rewriter incurs a penalty proportional to the number of rewritten queries, receiving a $-0.5$ penalty if more than four queries are generated. The Reranker is penalized based on the format and redundancy of selected documents, with a $-0.5$ penalty for duplicated entries or index out-of-bounds errors. The Answerer faces a penalty tied to the length of the generated answer, incurring a $-1.0$ penalty for excessively long responses. This reward structure facilitates precise control over individual agent behavior during optimization.

\section{Detailed Experiment Results}
\begin{figure*}[t]
    \centering
    \includegraphics[width=1\linewidth]{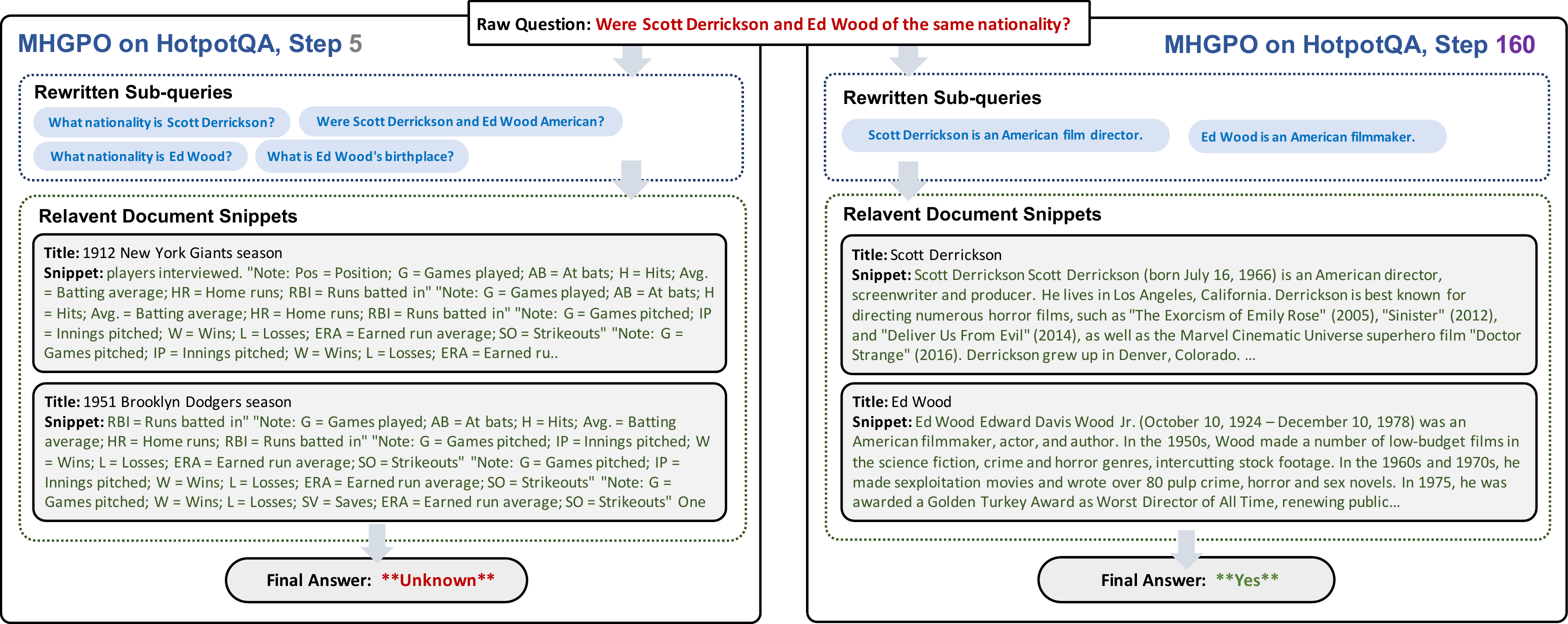}
    \caption{Case study: An example question is processed by the example \gls{mas}, with outputs from each stage shown for comparison between \gls{ours}-\gls{fof} after 5 steps and after 160 steps of training on HotpotQA.}
    \label{fig:example}
\end{figure*}

\begin{figure}
    \centering
\includegraphics[width=1\linewidth]{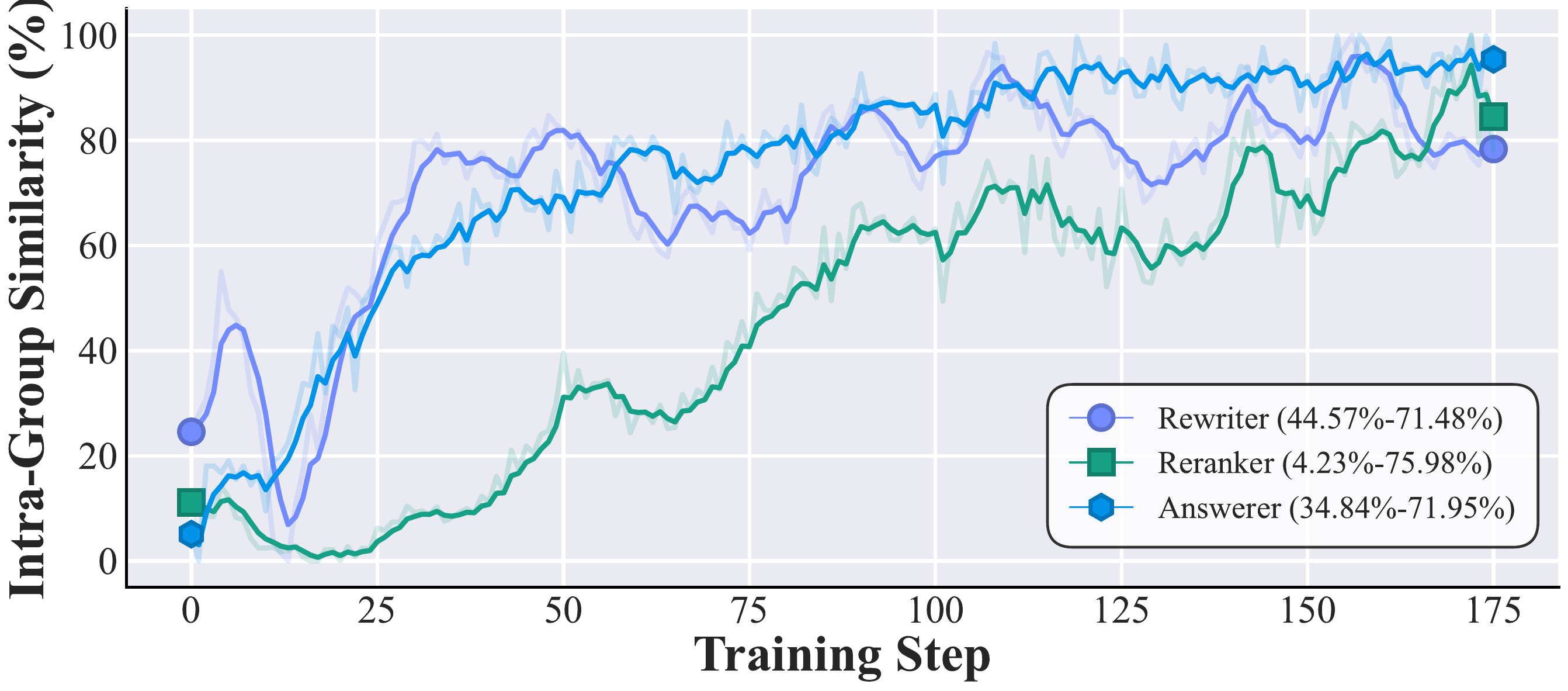}
    \caption{The intra-group similarity for each agent's rollout, measured by the average pairwise F1 score between generated responses. For consistency and ease of comparison across agents, similarity scores are normalized to the range [0, 1], with the corresponding minimum and maximum values indicated in the legend.}
    \label{fig:sim}
\end{figure}

\subsection{Training Dynamics}


In order to meticulously observe and compare the characteristics of different \gls{marl} algorithms during the training process, we have recorded the detailed metrics during the \gls{marl} training process on HotpotQA for the example \gls{mas} under consideration. Figure~\ref{fig:penalty} separately records the changes in the format penalties obtained by different agents during the training process within the considered example 3-agent \gls{mas}. It can be observed that all methods can quickly find the correct output format and thus avoid penalties (the decrease within 10 steps reaches within 0.1). During this process, \gls{mappo} once again demonstrates instability, with the possibility of the output format of agents collapsing midway. In addition, \gls{fof} and \gls{rr} are hardly troubled by the format and consistently obtain relatively low penalties.

\subsection{Impact of Group Size}

\begin{table}[H]
\caption{Examples of multi-hop questions and answers in HotpotQA.}
\label{tab:exp-hotpot}
\resizebox{\columnwidth}{!}{%
\begin{tabular}{c|c}
\hline
Question                                                                                                      & Answer \\ \hline
\begin{tabular}[c]{@{}c@{}}What is the relationship of Yeshahework \\ Yilma's mother to the man who was \\ Ethiopia's emperor from 1930 to 1974?\end{tabular} &
  niece \\ \hline
\begin{tabular}[c]{@{}c@{}}What role on "Switched at Birth" was played\\  by the actor who is being replaced by Daniel \\ Hall on "The Young and the Restless"?\end{tabular} &
  Tyler "Ty" Mendoza \\ \hline
\begin{tabular}[c]{@{}c@{}}Are the Sleepers located north or south \\ of the Kancamagus Highway?\end{tabular} & south  \\ \hline
\end{tabular}%
}
\end{table}

Table~\ref{tab:group} studies how the heterogeneous-group size in \gls{ours}-\gls{fof} affects final performance. Increasing the group size from 3 to 4 yields a clear jump across all metrics (ACC/EM/F1), indicating that a larger comparison set provides a stronger relative-reward signal for advantage estimation. Performance then largely saturates: group sizes 4--6 achieve very similar F1 (49.38--49.53) with small variance across seeds. The best overall average is obtained at group size 6 (AVG 42.284), which also attains the highest ACC and EM, while group size 4 achieves the top F1 and a comparable AVG. We therefore use $G=4$ as the default in the main experiments: it lies in this near-saturated regime and offers almost the same final performance as larger groups with lower rollout cost. In other words, $G=4$ is chosen as a compute--performance trade-off rather than as the single best setting on every metric.

\subsection{Progressive Homogenization of Rollout Groups}

\begin{figure}
    \centering
    \includegraphics[width=1\linewidth]{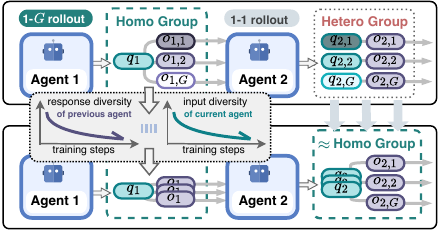}
    \caption{Illustration of how the heterogeneous groups produced by rollout sampling in \gls{mas} gradually transition into homogeneous groups during \gls{ours}.}
    \label{fig:diversity}
\end{figure}

We hypothesize that the effectiveness of Heterogeneous-group-based Advantage Estimation in \gls{ours} stems, in part, from the progressive convergence of heterogeneous groups into homogeneous ones during training: sampled trajectories within each group gradually become more uniform. This is supported by empirical evidence in Figure~\ref{fig:diversity} and Figure~\ref{fig:sim}. The intra-group similarity, quantified by the average pairwise F1 score, consistently increases for each agent throughout training. For the \emph{Rewriter}, output similarity reaches approximately 0.7 between steps 25 and 30, suggesting that the \emph{Reranker} may receive increasingly similar retrieved evidence. This, in turn, may make the \emph{Reranker}'s local inputs closer to a homogeneous group and could contribute to the performance improvements observed from step 25 onward.

\paragraph{Evidence consistent with task-specific convergence.} A natural concern is whether the increasing intra-group similarity signals entropy collapse---a degenerate state where the model loses the ability to generate diverse outputs. Our current evidence does not by itself rule out entropy collapse, but it is at least consistent with task-specific convergence in this setting. Unlike general-purpose LLM RL, where maintaining high entropy is crucial for broad exploration, agents in a \gls{mass} act as specialized sub-task solvers (Rewriter, Reranker, Answerer) with relatively narrow objectives. The observed homogenization coincides with each agent settling on high-reward behavioral patterns for its role: the Rewriter converges to generating two concise sub-queries, the Reranker to selecting 2--3 relevant documents, and the Answerer to producing succinct responses (Figure~\ref{fig:training-agent}). It also coincides with improved validation F1 during training (Figure~\ref{fig:training}) and stronger final QA performance (Table~\ref{tab:main}). Taken together, these observations suggest that the model is discovering higher-reward collaborative strategies rather than exhibiting an immediately harmful collapse.

\subsection{Case Study}

To investigate the specific improvements brought by \gls{marl} to multi-hop question answering, we examine the performance of the proposed \gls{mas} system on a representative QA example, comparing its behavior in the early stage (step 5) and the later stage (step 160) of reinforcement learning, as illustrated in Figure~\ref{fig:example}. 

At step 5, the system produces an incorrect answer (\textit{unknown}) because the selected relevant snippets do not contain useful information, despite the seemingly reasonable sub-queries generated by the Rewriter. In contrast, by step 160, the \gls{mas} successfully answers the question. This improvement stems from the Reranker correctly identifying two relevant documents---biographical entries on Scott Derrickson and Ed Wood. Interestingly, the Rewriter now produces only two sub-queries, which, while less intuitive to humans, align better with the behavior of the \texttt{contriever} retriever engine, enabling the retrieval of the desired documents. 

This example highlights how reinforcement learning can help agents like the Rewriter adapt to their environment (i.e., the retrieval engine), thereby overcoming performance bottlenecks.

\section{Single-Context Transcript Equivalence via Parameter Sharing and Reward Propagation}
\label{app:single_context_equiv}
\newtheorem{proposition}{Proposition}

In this section, we show that \gls{ours}'s \textbf{parameter sharing + reward propagation} yields an equivalent learning problem to training a single \gls{llm} that generates one \emph{single-context transcript} (e.g., $\langle q, a, \texttt{<toolcall>}, q, a, \ldots\rangle$): the per-token reward-weighted score terms match between the \gls{mas} execution and this single-transcript view. We detail the analysis below.


\subsection{MAS Trajectory and Single-Context Transcript}

Fix an input question $q\sim\mathcal{D}$ and a sampled trajectory index $i$.
The \gls{mas} executes agents and produces
\[
(q_{1,i}, o_{1,i}) \;\rightarrow\; (q_{2,i}, o_{2,i}) \;\rightarrow\; \cdots \;\rightarrow\; (q_{n,i}, o_{n,i}),
\]
where $q_{1,i}=q$ and $o_{k,i}=(o_{k,i,1},\dots,o_{k,i,T_{k,i}})$ is agent $k$'s output sequence.
Each agent input $q_{k,i}$ is constructed from system templates, tool outputs, and upstream realized texts.

By contrast, we construct a \emph{single-context transcript} that alternates fixed workflow text with generated tokens:
\begin{gather*}
x_{1,i} \triangleq q_{1,i},\\
x_{k+1,i} \triangleq \mathrm{Build}_{k+1}(x_{k,i}, o_{k,i}), \quad k=1,\dots,n,
\end{gather*}
where $x_{k,i}$ is the full prefix immediately before sampling $o_{k,i}$, and $\mathrm{Build}_{k+1}$ deterministically appends role templates, tool-call markers, tool outputs, separators, and required metadata. By construction, $q_{k,i}$ is a suffix of $x_{k,i}$.

Let the final system output $o_i$ be a deterministic function of $\{o_{k,i}\}_{k=1}^n$, with terminal reward $R_i$. We show below that \gls{ours} over an \gls{mas} rollout is equivalent to \gls{rl} over this single-context transcript.

\subsection{Assumptions}

\paragraph{(A1) Parameter sharing.}
All agents share the same backbone autoregressive policy $\pi_\theta$.

\paragraph{(A2) Deterministic prompt construction.}
All tool/retrieval results that influence downstream generation are explicitly \emph{materialized as tokens} in $x_{k+1,i}$. Consequently, in the single-context transcript, $\mathrm{Build}_{k+1}$ is deterministic given $(x_{k,i}, o_{k,i})$; equivalently, in the \gls{mas} formulation, $q_{k+1,i}=\Phi_k(o_{k,i})$ for a deterministic mapping $\Phi_k$.

\paragraph{(A3) Context sufficiency (inter-agent Markov property).}
For each $k,i$, the constructed prompt $q_{k,i}$ is sufficient for agent $k$'s generation:
\[
\pi_\theta(o_{k,i}\mid x_{k,i}) \;=\; \pi_\theta(o_{k,i}\mid q_{k,i}).
\]
This assumption is strong in general, but it is well aligned with our \gls{mas} setting.
Concretely, $q_{k,i}$ is produced by a deterministic prompt-construction routine that \emph{materializes} into text all upstream agent outputs and all tool-call returns that are available to the system (Assumption~A2), and the backbone LLM is always queried on $q_{k,i}$ (a suffix of $x_{k,i}$) with fixed role templates.
In our scenario, tool calls (retrieval, search, etc.) return discrete results (e.g., top-$K$ documents, snippets, or structured tool outputs) that are explicitly inserted into $q_{k,i}$; therefore, conditioning on the full transcript prefix $x_{k,i}$ provides no additional actionable information beyond what is already contained in $q_{k,i}$.
In other words, $q_{k,i}$ is an information-preserving summary of the preceding transcript with respect to agent $k$'s decision, which is precisely the intended design of the workflow prompt synthesis.

We note that A3 serves primarily as an idealized \emph{theoretical anchor} for establishing the single-context equivalence, rather than a strict operational prerequisite for \gls{ours}. Its validity depends on the \gls{mas}'s prompt construction mechanism rather than the partially observable nature of the underlying task: if the prompt is constructed to be strictly equivalent to a crafted linear transcript, A3 holds mathematically; if the \gls{mas} design ensures that each local prompt contains all decision-critical information---for instance, an Answerer relying solely on retrieved documents, independent of the Rewriter's exact query phrasing---A3 is practically satisfied. When practical prompts compress or hide history to manage context limits, A3 is relaxed; \gls{ours} is explicitly designed to handle this relaxed regime, and our experiments confirm that it achieves effective global optimization even when local contexts are imperfect (see Section~4).


\subsection{Probability Equivalence to a Single-Context Transcript}

\begin{proposition}[Single-context transcript probability equivalence]
\label{prop:single_context_prob_equiv}
Under Assumptions (A1)--(A3),
the \gls{mas} rollout distribution equals the distribution induced by a single autoregressive LLM generating the same transcript:
\begin{align*}
    \mathrm{MAS}_\theta(o_{1,i},\dots,o_{n,i}\mid q_{1,i})\\
\;=\;
\pi_\theta(o_{1,i},\dots,o_{n,i}\mid x_{1,i}),
\end{align*}
and both factorize as
\[
\prod_{k=1}^{n}\;\prod_{t=1}^{T_{k,i}}
\pi_\theta\!\left(o_{k,i,t}\mid x_{k,i}, o_{k,i,<t}\right).
\]
\end{proposition}

\begin{proof}
We show that both processes assign the same probability to each generated token, hence the same joint probability.

\paragraph{Single-transcript process.}
Immediately before generating $o_{k,i}$, the transcript prefix is $x_{k,i}$.
Autoregressive generation yields
\begin{gather*}
\pi_\theta(o_{1,i},\dots,o_{n,i}\mid x_{1,i}) \\
=\prod_{k=1}^{n}\;\prod_{t=1}^{T_{k,i}}
\pi_\theta(o_{k,i,t}\mid x_{k,i}, o_{k,i,<t}).
\label{eq:single_transcript_factor}
\end{gather*}


\paragraph{MASS process.}
At stage $k$, the system constructs the agent prompt $q_{k,i}$ and samples $o_{k,i}$ token-by-token from $\pi_\theta(\cdot\mid q_{k,i})$:
\[
\mathrm{MAS}_\theta(o_{k,i}\mid \text{history})
=
\prod_{t=1}^{T_{k,i}}
\pi_\theta(o_{k,i,t}\mid q_{k,i}, o_{k,i,<t}).
\]
By context sufficiency (A3), conditioning on $x_{k,i}$ is equivalent to conditioning on $q_{k,i}$ for generating $o_{k,i}$, hence
\begin{gather*}
    \pi_\theta(o_{k,i,t}\mid x_{k,i}, o_{k,i,<t}) \\
= \pi_\theta(o_{k,i,t}\mid q_{k,i}, o_{k,i,<t}),
\qquad \forall k,t.
\end{gather*}
Thus the \gls{mas} joint probability over all generated tokens equals \eqref{eq:single_transcript_factor}.
Deterministic insertion of workflow/tool tokens (A2) does not affect the probability mass on generated tokens,
so the two induced joint distributions coincide.
\end{proof}

\subsection{Reward Propagation and Token-Level Learning-Signal Equivalence}

We now formalize the role of \textbf{reward propagation} in \gls{ours} and show that,
together with parameter sharing, it establishes \emph{token-level learning-signal equivalence}
between the \gls{mas} execution and a single-LLM transcript rollout---not only for \textbf{score-function gradients},
but also for \textbf{GRPO-style objectives} with group-based advantages.

\paragraph{Reward broadcasting in \gls{ours}.}
In MHGPO, the terminal reward $R_i$ is first propagated backward to all agents in the trajectory
(Eq.~\ref{eq:aggr}), and is further \emph{broadcast to all generated tokens} within each agent output,
exactly as in GRPO.
Concretely, for each trajectory $i$, every generated token $o_{k,i,t}$ is optimized with respect to
the same scalar learning signal (up to normalization by group statistics).
This design ensures that all token decisions---regardless of which agent produced them---are treated as
contributing to the final system outcome.

\subsubsection{Token-Level Equivalence for Score-Function Policy Gradients}

We begin by proving token-level learning-signal equivalence under the naive score-function (REINFORCE) policy-gradient estimator.

\begin{proposition}[Token-level learning-signal equivalence (score-function)]
\label{prop:token_pg_equiv}
Under Proposition~\ref{prop:single_context_prob_equiv}, and assuming the terminal reward $R_i$
is broadcast to all generated tokens in trajectory $i$, the score-function gradient computed from
MASS rollouts equals that computed from the single-LLM transcript rollout:
\begin{gather*}
\nabla_\theta J(\theta)
= \\
\mathbb{E}\!\left[
R_i \sum_{k=1}^{n}\sum_{t=1}^{T_{k,i}}
\nabla_\theta \log \pi_\theta(o_{k,i,t}\mid x_{k,i},o_{k,i,<t})
\right].    
\end{gather*}

Each summand corresponds to the same token under the same conditional distribution in both views.
\end{proposition}

\begin{proof}
By Proposition~\ref{prop:single_context_prob_equiv}, the joint likelihood of the MASS rollout
equals the likelihood of the single-context transcript.
Broadcasting $R_i$ to all generated tokens yields a score-function gradient that is a sum over
per-token log-probabilities weighted by the same terminal reward.
Since each token has the same conditional distribution in both views, the per-token learning signal
is identical.
\end{proof}

\subsubsection{Token-Level Equivalence for GRPO-style Policy Gradients}
\label{app:token_grpo_equiv}

We now formalize the token-level equivalence of \gls{ours}-\gls{fof} under GRPO-style policy gradients.
In \gls{fof}, each agent's advantage is computed over \emph{heterogeneous groups}: for a fixed raw input, a group collects multiple generations of the same agent across different \gls{mas} trajectories originated from the same system input. With this construction, \textbf{parameter sharing + reward propagation + token-level reward broadcasting} yields an exact token-wise correspondence to standard GRPO on a single LLM, executed over a single-context transcript that interleaves fixed workflow/tool tokens with generated tokens.

\paragraph{Importance sampling ratio equivalence.}
For any generated token $o_{k,i,t}$, the token-level importance ratio in \gls{grpo} is
\[
r_{k,i,t}(\theta)
\;\triangleq\;
\frac{\pi_\theta\!\big(o_{k,i,t}\mid x_{k,i},o_{k,i,<t}\big)}
{\pi_{\theta_{\text{old}}}\!\big(o_{k,i,t}\mid x_{k,i},o_{k,i,<t}\big)}.
\]
By Proposition~\ref{prop:single_context_prob_equiv}, the conditional distribution of each token given its single-context prefix $x_{k,i}$
is the same under both the \gls{mas} rollout and the single-transcript execution. Hence the ratio $r_{k,i,t}(\theta)$ is identical in the two views.

\paragraph{Advantage equivalence.}
In \gls{ours}-\gls{fof}, grouping is \emph{per-agent under a fixed system input}: for a fixed query $q$ and each agent $k$, $\mathcal{G}_k(q)$ collects the $G$ rollouts of agent $k$ induced by the $G$ trajectories originated from $q$. This is \emph{heterogeneous} at the agent level for $k>1$, because each agent prompt $q_{k,i}$ is a deterministic function of rollout-specific upstream text (while all share the same root $q$).
Let $R_{k,i}$ be the propagated reward for agent $k$ in rollout $i$ (Eq.~(\ref{eq:aggr})). \gls{ours}-\gls{fof} normalizes $\{R_{k,j}\}_{j\in\mathcal{G}_k(q)}$ \emph{within each agent's group} to obtain a scalar advantage $A_{k,i}$, and broadcasts it to every token in $o_{k,i}$.

By Proposition~\ref{prop:single_context_prob_equiv}, each $(k,i)$ corresponds to a transcript segment within a trajectory sampled under $\pi_\theta$ for the same $q$. Consider the single-transcript counterpart in which \gls{grpo} is applied with the same per-agent grouping---that is, advantages are computed by normalizing rewards of the same transcript segment (agent role) across the $G$ transcripts rooted at $q$. Under this construction, the normalization set $\{R_{k,j}\}_{j\in\mathcal{G}_k(q)}$ coincides in the two views, yielding identical token-wise advantages. If agent-specific rewards are included, they map directly to step-wise rewards on the same transcript MDP, and the same correspondence holds.

\begin{proposition}[Token-level GRPO equivalence for MHGPO-FoF]
\label{prop:grpo_token_equiv_fof}
Under Assumptions (A1)--(A3), with reward propagation and token-level broadcasting as in MHGPO-FoF, and with per-agent grouping under a fixed system input, the GRPO surrogate yields identical token-wise learning signals in the \gls{mas} view and the single-transcript view.
In particular, for every generated token $o_{k,i,t}$,
\begin{equation}
r^{\text{MAS}}_{k,i,t}(\theta)=r^{\text{tr}}_{k,i,t}(\theta),
\qquad
A^{\text{MAS}}_{k,i,t}=A^{\text{tr}}_{k,i,t}=A_{k,i},
\label{eq:fof_adv_equal}
\end{equation}
where $A_{k,i}$ is the per-agent FoF group-relative advantage computed by normalizing $\{R_{k,j}\}_{j\in\mathcal{G}_k(q)}$ and broadcast to all tokens of $o_{k,i}$.
Hence the clipped GRPO term is identical token-by-token in both views.
\end{proposition}

\begin{proof}
By Proposition~\ref{prop:single_context_prob_equiv}, each token has the same conditional distribution given the same transcript prefix $x_{k,i}$
in both executions, so $r^{\text{MAS}}_{k,i,t}(\theta)=r^{\text{tr}}_{k,i,t}(\theta)$.
Under per-agent grouping, both views (i) normalize the \emph{same} propagated rewards $\{R_{k,j}\}_{j\in\mathcal{G}_k(q)}$ and (ii) broadcast the resulting scalar $A_{k,i}$ to every token of $o_{k,i}$,
giving \eqref{eq:fof_adv_equal}. Since clipping is deterministic in $(r_{k,i,t}(\theta),A_{k,i})$, the per-token GRPO surrogate matches.
\end{proof}

\subsection{Summary and key takeaway.}
Under Assumptions (A1)--(A3), the token-level learning signal of \gls{ours} can be interpreted as standard GRPO applied to a \emph{constructed single-context transcript} sampled from a single policy $\pi_\theta$, where workflow/tool-call tokens are explicitly interleaved with generated tokens (akin to agentic traces in systems such as R1-Searcher). Concretely, reward back-propagation followed by GRPO-style broadcasting assigns each generated token a \emph{group-relative} advantage target that matches the GRPO advantage computed on the corresponding transcript trajectories.

This equivalence is primarily a \emph{conceptual} bridge rather than an implementation claim: \gls{ours} still performs \emph{multi-task, multi-context} optimization, which empirically improves convergence and stability over naively applying single-context GRPO to long transcripts. Assumption (A3) (\emph{Context Sufficiency}) serves as the theoretical anchor for this bridge, formalizing the gap between the idealized single-transcript view and practical multi-context execution. \textbf{When A3 holds, \gls{ours} yields token-level alignment with GRPO; when A3 is relaxed---as is typical in practical search tasks where agents condition on compressed or incomplete local contexts---\gls{ours} exploits implicit inter-agent correlations through heterogeneous grouping to achieve global coordination that single-context methods cannot.}

The key to this robustness lies in the design philosophy of heterogeneous grouping: by comparing rollouts induced by the same system input but with different intermediate decisions, it intentionally shifts the optimization focus from local agent performance to global system success. A downstream agent conditioned on a poor upstream prefix may receive low rewards regardless of its own actions; rather than treating this as noise, heterogeneous grouping \emph{implicitly down-weights} such coordination-failure trajectories and steers the joint policy toward globally successful interactions. The Round-Robin strategy further stabilizes training by serving as a bridge between global coordination pressure and local learning stability, stochastically mixing heterogeneous updates with homogeneous baselines to provide each agent with both system-level signals and clean, input-conditioned comparisons. A formal variance analysis of heterogeneous grouping under partial observability remains an important direction for future work.




\begin{algorithm}[htbp]
\SetAlgoLined
\DontPrintSemicolon
\KwIn{Question $q$, rollout group size $G$, shared actor parameters $\theta$}
\KwOut{Set of rollout trajectories $\{\mathcal{T}_i\}_{i=1}^G$}

\tcp{Determine the first agent to handle the question}
$A_{c_0} \gets \gls{mas}.next\_agent(q)$\;

\tcp{Call fork\_on to perform Fork-on-First rollout}
$\{\mathcal{T}_i\}_{i=1}^G \gets \texttt{fork\_on}(q, G, A_{c_0}, \theta)$\;

\Return $\{\mathcal{T}_i\}_{i=1}^G$
\caption{Fork-on-First Rollout Sampling}
\label{algo:fof}
\end{algorithm}

\begin{algorithm}[htbp]
\SetAlgoLined
\DontPrintSemicolon
\KwIn{Single question $q$, rollout group size $G$, shared actor parameters $\theta$}
\KwOut{Merged set of rollout pairs $\mathcal{P}$ collected from each individual agent}

Initialize $\mathcal{P} \gets \emptyset$\;

\For{each agent $A_i$ where $i = 1$ to $n$}{
    \tcp{Fork on agent $A_i$}
    Extract rollout pairs $\{(q_{k,j}, o_{k,j}, m_{k,j})\}$ from $\texttt{fork\_on}(q, G, A_i, \theta)$\;

    \tcp{Filter only rollout pairs where $k = i$}
    \For{each rollout pair $(q_{k,j}, o_{k,j}, m_{k,j})$}{
        \If{$k = i$}{
            Append $(q_{k,j}, o_{k,j}, m_{k,j})$ to $\mathcal{P}$\;
        }
    }
}

\Return final rollout pairs $\mathcal{P}$
\caption{Independent Sampling}
\label{algo:is}
\end{algorithm}

\begin{algorithm}[htbp]
\SetAlgoLined
\DontPrintSemicolon
\KwIn{Batch of questions $\mathcal{B} = \{q_1, q_2, \ldots, q_B\}$, rollout group size $G$, shared actor parameters $\theta$, agent sampling probabilities $\{p_1, p_2, \ldots, p_n\}$}
\KwOut{Set of rollout trajectories $\{\mathcal{T}_i\}_{i=1}^{G'}$ with consistent group identifiers}

Initialize rollout pair set $\mathcal{P}_{\text{all}} \gets \emptyset$\;

\For{each question $q \in \mathcal{B}$}{
    \tcp{Sample fork agent from categorical distribution}
    $A_i \sim \text{Categorical}(p_1, p_2, \ldots, p_n)$\;

    \tcp{Fork-on-$A_i$ to obtain $G$ trajectories}
    $\{\mathcal{T}^{(q)}_j\}_{j=1}^G \gets \texttt{fork\_on}(q, G, A_i, \theta)$\;

    \For{each trajectory $\mathcal{T}^{(q)}_j$}{
        Extract all rollout pairs $(q_{k,j}, o_{k,j}, m_{k,j})$ from $\mathcal{T}^{(q)}_j$ and append to $\mathcal{P}_{\text{all}}$\;
    }
}

\tcp{Group reassignment for rollout pairs with group size $=1$}
Count the number of occurrences for each group identifier $m$ in $\mathcal{P}_{\text{all}}$\;

Let $\mathcal{P}_{\text{single}} \gets$ rollout pairs whose original group $m$ appears only once\;

Let $\mathcal{P}_{\text{valid}} \gets$ rollout pairs whose group $m$ appears more than once\;

Shuffle $\mathcal{P}_{\text{single}}$ randomly\;

Partition $\mathcal{P}_{\text{single}}$ into $N = \lfloor |\mathcal{P}_{\text{single}}| / G \rfloor$ disjoint groups $\{g_i\}_{i=1}^{N}$, where each $g_i = \{(q_{i,j}, o_{i,j})\}_{j=1}^G$\;

\For{each group $g_i$ where $i = 1, \ldots, N$}{
    \For{each rollout pair $(q_{i,j}, o_{i,j})$ where $j = 1, \ldots, G$}{
        Assign new group identifier: $m_{i,j} \gets i + |\mathcal{P}_{\text{valid}}|$\;
    }
}

Merge $\mathcal{P}_{\text{valid}}$ with all reassigned $g_i$ groups to form final $\mathcal{P}_{\text{all}}$\;

\Return final rollout pairs $\mathcal{P}_{\text{all}}$
\caption{Round-Robin Rollout Sampling}
\label{algo:rr}
\end{algorithm}

\begin{algorithm}[htbp]
\SetAlgoLined
\DontPrintSemicolon
\KwIn{Training dataset $\mathcal{D}$ with question-answer pairs $(q, a_{\text{golden}})$, reward function $\mathcal{R}$, \texttt{batch\_size}, \texttt{total\_epochs}, \texttt{ppo\_epochs}, rollout sampling function $\mathcal{H}$}
\KwOut{Optimized actor parameters $\theta$}
Initialize actor parameters $\theta$ (shared among all agents), reference parameters $\theta_{\text{ref}}$\;

\For{epoch $= 1$ \KwTo \texttt{total\_epochs}}{
    \For{each batch $\mathcal{B} \subset \mathcal{D}$}{
        \For{each question $q$ in batch $\mathcal{B}$}{
            \tcp{Multi-Agent rollout trajectory sampling}
            Generate $G$ rollout trajectories $\{\mathcal{T}_i\}_{i=1}^{G} \gets \mathcal{H}(\theta, q)$\;
            Collect rollout pairs $\{(q_{k,j}, o_{k,j},m_{k,j})\}$ for all agents across all trajectories\;

            \tcp{Final reward calculation}
            \For{each trajectory $\mathcal{T}_i$}{
                Let $o_i$ be the final output of $\mathcal{T}_i$\;
                Compute final reward: $R_i^{\text{shared}} = \mathcal{R}(o_i, a_{\text{golden}})$\;
            }

            \tcp{Backward propagation of rewards}
            \For{each rollout pair $(q_{k,j}, o_{k,j})$}{
                Calculate aggregated reward $R^\text{shared}_{k,j}$ according to Equation \ref{eq:aggr}\;
                Apply agent-specific reward $R^\text{spe}_{k,j}$ to obtain $R_{k,j}$\;
            }
        }

        \tcp{Actor update}
        \For{ppo\_epoch $= 1$ \KwTo \texttt{ppo\_epochs}}{
            \For{each rollout pair $(q_{k,j}, o_{k,j}, m_{k,j}, R_{k,j})$}{
                Form rollout groups based on $m_{k,j}$ and estimate the advantage using Equation \ref{eq:adv}\;
            }
            Update policy parameters $\theta$ using the policy gradient method with Equation \ref{eq:target} as the target\;
        }
    }
}
\caption{Multi-Agent Optimization in MHGPO}
\label{algo:ours}
\end{algorithm}
\FloatBarrier
\begin{algorithm}[htbp]
\SetAlgoLined
\DontPrintSemicolon
\KwIn{Question $q$, rollout group size $G$, fork agent $A_i$, shared actor parameters $\theta$}
\KwOut{Set of rollout trajectories $\{\mathcal{T}_i\}_{i=1}^G$}

Initialize $o \gets q$, empty prefix trajectory $\mathcal{T}_{\text{prefix}} \gets \emptyset$

\tcp{One-to-one rollout until fork agent $A_i$ is encountered}
\While{$\gls{mas}.next\_agent(o) \neq A_i$}{
    $A_{c_j} \gets \gls{mas}.next\_agent(o)$\;
    
    $q_{c_j} \gets \gls{mas}.process\_prompt(o, A_{c_j})$\;
    
    $o_{c_j} \gets A_{c_j}(q_{c_j}; \theta)$\;
    
    Append $(q_{c_j}, o_{c_j})$ to $\mathcal{T}_{\text{prefix}}$\;
    
    $o \gets o_{c_j}$\;
}

\tcp{Fork-on-$A_i$: one-to-many rollout from agent $A_i$}
$q_{c_i} \gets \gls{mas}.process\_prompt(o, A_i)$\;

\For{$i = 1$ \KwTo $G$}{
    $o_{c_i,i} \gets A_i(q_{c_i}; \theta)$\;

    Initialize $\mathcal{T}_i \gets \mathcal{T}_{\text{prefix}} \cup \{(q_{c_i}, o_{c_i,i})\}$\;

    $o \gets o_{c_i,i}$\;

    \tcp{Continue one-to-one rollout for each fork}

    $j \gets 1$ \;
    
    \While{$\gls{mas}.has\_next\_agent(o)$}{
            $A_{c_{i+j}} \gets \gls{mas}.next\_agent(o)$\;

            $q_{c_{i+j}} \gets \gls{mas}.process\_prompt(o, A_{c_{i+j}})$\;

            $o_{c_{i+j}} \gets A_{c_{i+j}}(q_{c_{i+j}}; \theta)$\;

            Append $(q_{c_{i+j}}, o_{c_{i+j}})$ to $\mathcal{T}_i$\;

            $o \gets o_{c_{i+j}}$\;
            
            $j\gets j+1$
        }
    }

\tcp{Group identifier assignment}
\For{$i = 1$ \KwTo $G$}{
    \For{each rollout pair $(q_{i,*}, o_{i,*})$ in trajectory $\mathcal{T}_i$}{
        Assign group identifier $m_{i,*} \gets i$
    }
}

\Return $\{\mathcal{T}_i\}_{i=1}^G$
\caption{Base Rollout Sampling Function \textit{fork\_on()}}
\label{algo:fo}
\end{algorithm}


\end{document}